\documentclass{article}

\PassOptionsToPackage{numbers, compress}{natbib}
\usepackage[preprint]{neurips_2025}

\usepackage[utf8]{inputenc} 
\usepackage[T1]{fontenc}    
\usepackage{hyperref}       
\usepackage{url}            
\usepackage{booktabs}       
\usepackage{amsfonts}       
\usepackage{nicefrac}       
\usepackage{microtype}      
\usepackage{xcolor}         
\usepackage{graphicx}
\usepackage{subcaption}
\usepackage{algorithm}
\usepackage{algpseudocode}
\usepackage{amsmath}
\usepackage{makecell}
\usepackage{multirow}
\usepackage{array}
\usepackage{bm}
\usepackage{colortbl}
\usepackage{wrapfig}
\definecolor{linkcolor}{RGB}{0,102,204}
\newcommand{\OURMODEL}{\texttt{ContentV}}
\newcommand{\eqrefe}[1]{Equation~\eqref{#1}}
\newcommand{\figrefe}[1]{Figure~\ref{#1}}
\newcommand{\tabrefe}[1]{Table~\ref{#1}}

\renewcommand{\paragraph}[1]{\vspace{0.05in}\noindent\textbf{#1.}\,\,}

\title{ContentV: Efficient Training of Video Generation Models with Limited Compute}
\author{
  Wenfeng Lin$^*$, Renjie Chen$^*$, Boyuan Liu, Shiyue Yan, Ruoyu Feng, Jiangchuan Wei, \\ \bf Yichen Zhang, Yimeng Zhou, Chao Feng, Jiao Ran, Qi Wu, Zuotao Liu, Mingyu Guo$^\dagger$ \\
  ByteDance Douyin Content Group \\
  $^*$ Equal contribution \quad $^{\dagger}$ Project leader
}
\begin{document}

\maketitle

\begin{figure}[htbp]
    \centering
    \includegraphics[width=\textwidth]{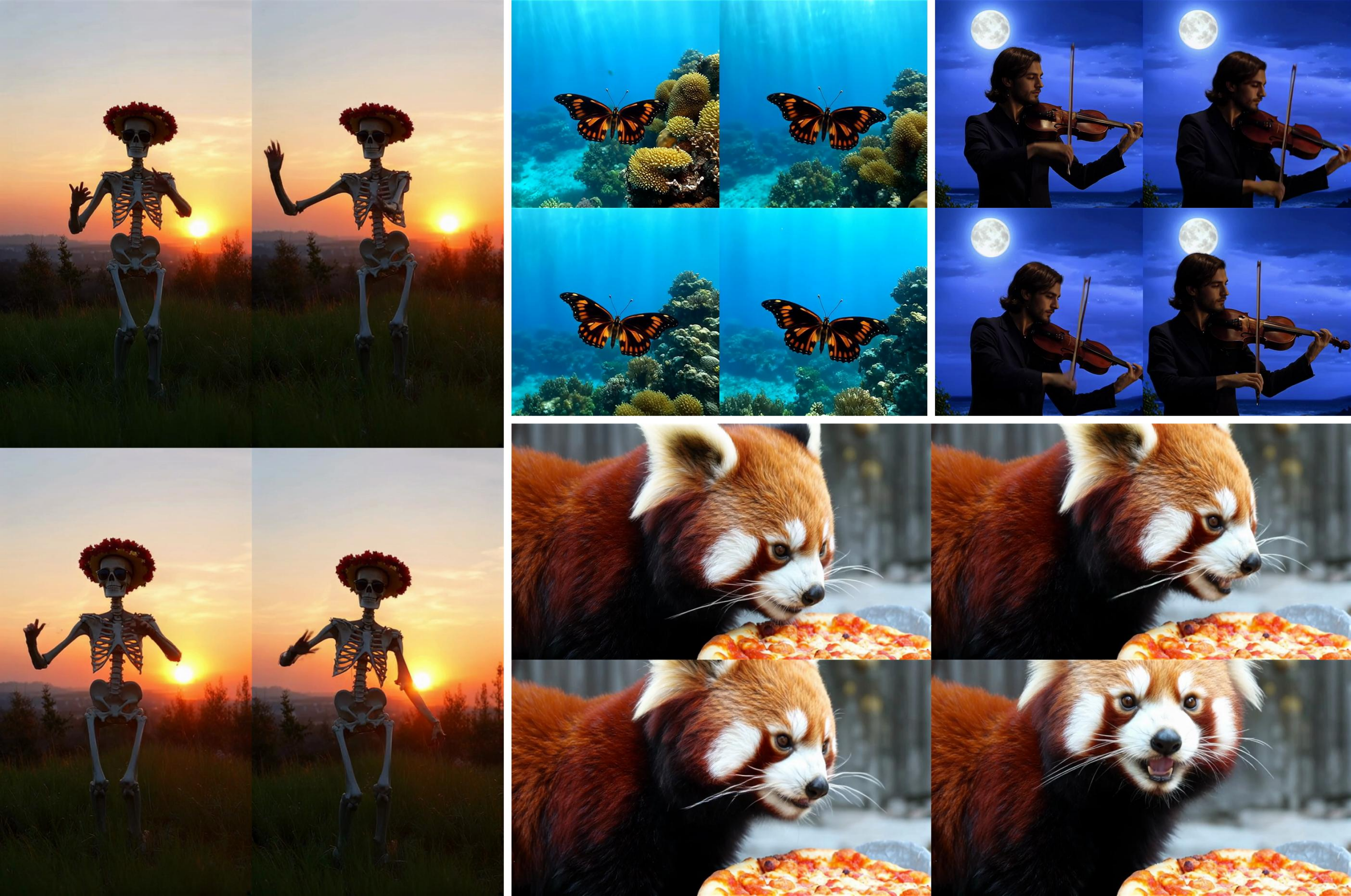}
    \caption{Video Generation Samples. \OURMODEL~can generate diverse, creative, and high-quality multi-resolution videos based on textual prompts.}
    \label{fig:example}
\end{figure}

\begin{abstract}
Recent advances in video generation demand increasingly efficient training recipes to mitigate escalating computational costs. In this report, we present \OURMODEL, an 8B-parameter text-to-video model that achieves state-of-the-art performance (\textbf{85.14} on VBench) after training on 256$\times$64~GB Neural Processing Units (NPUs) for merely four weeks. \OURMODEL~generates diverse, high-quality videos across multiple resolutions and durations from text prompts, enabled by three key innovations:
(1) A minimalist architecture that maximizes reuse of pre-trained image generation models for video generation;
(2) A systematic multi-stage training strategy leveraging flow matching for enhanced efficiency; and
(3) A cost-effective reinforcement learning with human feedback framework that improves generation quality without requiring additional human annotations.
All the code and models are available at: \color{linkcolor}\url{https://contentv.github.io}.
\end{abstract}

\section{Introduction}
Video generation has long been considered an extremely challenging task, primarily due to the difficulties in modeling complex temporal dynamics and spatial details. With the release of Sora~\cite{sora} by OpenAI, latent diffusion models based on the diffusion transformer (DiT) architecture have demonstrated significant potential in this field, spurring a series of open-source efforts such as Open-Sora-Plan~\cite{lin2024open}, CogVideoX~\cite{yang2024cogvideox}, HunyuanVideo~\cite{kong2024hunyuanvideo}, Step-Video-T2V~\cite{ma2025step}, and Wan2.1~\cite{wang2025wan}. 

However, current video generation models typically require massive computational resources for training. Moreover, the long-sequence nature of videos leads to exceptionally high GPU memory demands, necessitating the use of advanced GPUs, which severely hinders progress in this domain. For example, MovieGen~\cite{polyak2410movie} employed 6,144 H100 GPUs to train a 30B model on $\mathcal{O}$ (100~M) videos and $\mathcal{O}$(1~B) images. Additionally, we observe that existing methods predominantly adopt a two-stage training paradigm of "image-first, video-later." This raises a critical question: Can we directly leverage well-established image generation models from the open-source community (e.g., Stable Diffusion 3~\cite{sd3}, FLUX~\cite{Flux}) and adapt them to rapidly acquire video generation capabilities?

To address this question and the associated challenges, this technical report presents \OURMODEL, the first open-source video generation model based on Stable Diffusion 3.5 Large (SD3.5L) that is trained entirely on NPUs. We systematically share our data curation pipeline, model architecture adaptation, training infrastructure, as well as large-scale pre-training and post-training recipes. Our findings reveal that latent diffusion models exhibit remarkable adaptability to variational autoencoder (VAE) substitution. Specifically, by simply replacing the VAE in SD3.5L with a 3D-VAE and incorporating 3D positional encoding, the original image model can rapidly acquire video generation capabilities.
Through large-scale pre-training at multiple resolutions and durations, the model learns to generate diverse and creative videos conditioned on prompts. Further enhancements via supervised fine-tuning (SFT) and reinforcement learning with human feedback (RLHF) significantly improve its instruction-following ability and visual quality.

For training infrastructure, we designed a distributed training framework leveraging 64GB-memory NPUs (AI accelerators as GPU alternatives). By decoupling feature extraction and model training into separate clusters and integrating asynchronous data pipelines with 3D parallelism strategies, we achieved efficient training for 480P resolution, 24 FPS, 5-second videos. After one month of training on 256 NPUs, \OURMODEL~scored 85.14 on the VBench~\cite{vbench}, outperforming CogVideoX-5B~\cite{yang2024cogvideox} and HunyuanVideo-13B~\cite{kong2024hunyuanvideo}, while also maintaining a slight edge over Wan2.1-14B~\cite{wang2025wan} in human evaluations. This work establishes a new technical pathway for video generation research in resource-limited environments.

\section{Related Work}
In the past few years, diffusion models~\cite{ho2020denoising,dhariwal2021diffusion} have demonstrated exceptional capabilities in image generation~\cite{saharia2022photorealistic,rombach2022high,perniaswurstchen,sd3,Flux}. With the rapid advancement and widespread adoption of these image generation technologies, attention has increasingly shifted toward extending image into the temporal domain: video composed of multiple continuous frames. 

Early text-to-video generative models~\cite{singermake,ho2022imagen,guoanimatediff,gupta2023photorealistic,blattmann2023align,wang2024microcinema} primarily focused on adapting pre-trained text-to-image architectures for video generation by incorporating learnable temporal layers. These approaches typically employed hierarchical cascaded pipelines comprising distinct stages for keyframe generation, temporal upsampling, and spatial upsampling. However, several limitations hindered their effectiveness. Some methods \cite{singermake,ho2022imagen} performed video generation directly in pixel space, which proved computationally expensive and difficult to optimize. Furthermore, even when utilizing variational autoencoders to map video in pixel space to latent space, these approaches often failed to leverage the temporal redundancy inherent in video data, resulting in inefficient optimization and suboptimal performance. Additionally, the cascaded structure necessitated complex generation workflows and precluded end-to-end optimization.

The debut of Sora \cite{videoworldsimulators2024} in early 2024 represented a pivotal advancement in this field. Leveraging a 3D-VAE architecture with full attention mechanisms \cite{peebles2023scalable}, Sora demonstrated how scaling computational resources and training data could yield substantial improvements. Specifically, Sora achieved remarkable spatial-temporal consistency, generated extended video sequences with coherent narratives, and facilitated seamless video continuation. Furthermore, Sora's capacity to simulate physical interactions suggested its potential as a world model capable of encoding physical dynamics, thereby catalyzing intensified research interest in high-fidelity video synthesis methodologies.

Sora's impressive performance and substantial potential have sparked a research surge in academia and prompted increased resource investment from industry, driving rapid advancements in the field. Open-source models such as Open-Sora \cite{opensora}, Open-Sora-Plan \cite{lin2024open}, Mochi-1 \cite{genmo2024mochi}, CogVideoX \cite{yang2024cogvideox}, HunyuanVideo \cite{kong2024hunyuanvideo} and Wan2.1 \cite{wang2025wan} have provided a solid foundation for academic research, while commercial closed-source models like Kling \cite{kling}, Hailuo \cite{hailuo}, Vidu \cite{vidu}, PixelVerse \cite{pixelverse}, Runway Gen-3 \cite{gen3alpha}, Pika \cite{pika}, and Veo \cite{veo2} have established more user-friendly interfaces, facilitating widespread adoption and dissemination of video generation technology among the general public.

\section{Data Curation Pipeline}
In modern model training, data quality and diversity take precedence over quantity. Especially for generation tasks, low-quality data can deteriorate the generation results of the model. In this section, we describe the pipelines for large-scale high-quality data collection, and processing, as well as various processors and criteria for data filtering.

\subsection{Processing and Filtering}
We collected raw data from a diverse range of sources, including publicly available academic datasets~\cite{schuhmann2022laion, chen2024panda, wang2023internvid} and online CC0-licensed websites, such as Mixkit and Pexels. These datasets cover a broad spectrum of scenes, motions, and visual styles, which are crucial for training a video generation model with strong generalization capabilities.

Given the diverse sources of our raw data, a substantial portion of the videos are of low quality, which significantly degrades the performance of video generation. To address this, we use a comprehensive content-aware data filtering pipeline that operates on intrinsic video dimensions to enhance the overall quality of the dataset. 

\paragraph{Video Segmentation} The duration of the original videos ranges from a few seconds to several hours. Training requires splitting long videos into second-level video clips. Meanwhile, in order to avoid scene change in a video clip, we use PySceneDetect~\cite{PySceneDetect} to distinguish scene changes, cut long videos into continuous shots, and then split them into short video clips ranging from 3 to 6 seconds based on the duration. To ensure clear temporal boundaries, we further remove transitions with gradual scene changes using temporal pattern analysis.

\paragraph{Video Deduplication} We propose a hierarchical deduplication approach to address data redundancy and imbalance in video datasets. Specifically, we use an internal embedding model for video feature extraction. 
For the original long videos, we perform k-means clustering to group semantically similar content, and apply adaptive thresholding where clusters with higher density undergo more aggressive deduplication using stricter similarity thresholds. Video clips derived from the same source video often exhibit substantial content and scene overlap. To mitigate redundancy, we compute pairwise feature similarities and discard highly similar clips.

\paragraph{Subtitle and Watermark Detection} We utilize PaddleOCR~\cite{paddleocr2020} to locate and crop subtitles, as well as to remove video clips with excessive text. Furthermore, we also use an internal detection model to detect and remove sensitive information such as watermarks, logos, and black borders,.

\paragraph{Blur Detection} We detect blurriness by computing the variance of the Laplacian operator~\cite{pech2000diatom}, which serves as a quantitative measure of frame sharpness. Lower variance values correspond to increased blurriness, typically resulting from camera motion or focus inaccuracies.

\paragraph{Motion Dynamics} The dynamics of the main object in the video mainly come from two aspects: one is the movement of the object itself, and the other is the simultaneous movement of the foreground and background brought about by the movement of the camera lens. We take the average dynamic score of the foreground and the background as the overall dynamic score. Specifically, The dense optical flow is obtained by using GMFlow~\cite{xu2022gmflow}. The approximate background optical flow is obtained by fitting the perspective transformation, and then the foreground optical flow is rectified. So as to separate the foreground and background and calculate the dynamic scores respectively. 

\paragraph{Aesthetic Score} We use an improved aesthetic score predictor based on the CLIP model trained on LAION~\cite{schuhmann2022laion} to predict the aesthetic scores of key frames in each clip and calculate their average.

\begin{figure}[htbp]
    \centering
    \includegraphics[width=0.95\linewidth]{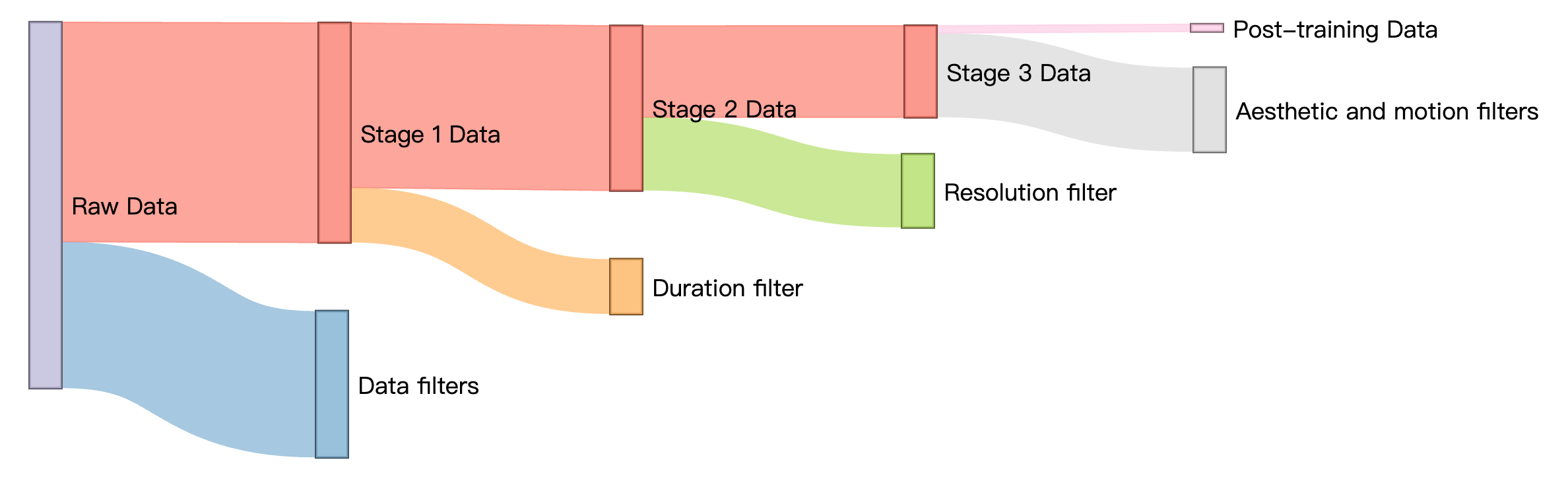}
    \caption{Overall of data filtering pipeline. For the filtered pre-training data, data from different training stages will be filtered out based on duration and resolution.}
    \label{fig:datapipeline}
\end{figure}

As shown in~\figrefe{fig:datapipeline}, using our data curation pipeline, we construct a high-quality visual dataset at a scale of $\mathcal{O}$(100~M) images, and $\mathcal{O}$(10~M) video clips, the duration range from 2 to 8 seconds. 

\subsection{Post-training Data Selection}
\label{sec:post-data}
After passing through the data filtering pipeline, a large number of low-quality and repetitive videos in the original data were filtered out, and a pre-training dataset of tens of millions of video clips was constructed through video slicing. However, due to the large quantity of pre-trained video data, the variation range of the quality of pre-trained data is very wide. Furthermore, we applied stricter filtering criteria on the pre-training data, obtaining a high-quality subset of one million. In particular, we emphasize the following refined selection criteria:
\begin{itemize}
    \item \textbf{Aesthetic Score.} The aesthetic model trained on laion is mainly for images and has a strong preference for high color contrast. And unlike a single image, the frames extracted from the video will have an inter-frame smoothing effect due to the need for temporal continuity. We trained an aesthetic model of video based on the VideoCLIP~\cite{wang2024videoclipxladvancinglongdescription}, which can consider both the visual quality of the picture and the degree of text alignment in the video simultaneously.
    \item \textbf{Motion Dynamics.} At this stage, we emphasize more on the motion dynamics of the main subject of the video. We calculate the dynamic score in a way similar to the foreground and background in the pre-training stage respectively. However, we give a higher weight to the foreground dynamic score and take the weighted dynamic score of the foreground and background as the total dynamic score. 
\end{itemize}
For the pre-training data, we use these two stricter criteria to score and rank the clips, and select the clips that are in the top 10\% in both criteria. A total of one million clips were obtained.

\subsection{Captioning}
As demonstrated by DALL-E 3~\cite{dalle3}, the precision and comprehensiveness of captions are essential for enhancing the prompt-following ability and output quality of generative models. 
We extract video frames at 1 fps and generate dense captions using the Qwen 2.5 VL~\cite{bai2025qwen2} multimodal model. We found that larger models yield more accurate captions while mitigating hallucination effects. However, generating dense captions for tens of millions of videos with the largest 72B model incurs prohibitively high computational costs. Therefore, we employ the 7B model for pre-training data, reserving the 72B model exclusively for the higher-quality subsets used in post-training.

\section{Architecture}
In this section, we present the architecture of \OURMODEL. Unlike previous approaches~\cite{yang2024cogvideox, wang2025wan, kong2024hunyuanvideo, moviegen, ma2025step} that prioritize novel structural designs, our methodology emphasizes the efficient adaptation of open-source text-to-image models, achieving video generation capability through minimal architectural modifications.

\subsection{Overview}
\begin{figure}[htbp]
    \centering
    \includegraphics[width=\textwidth]{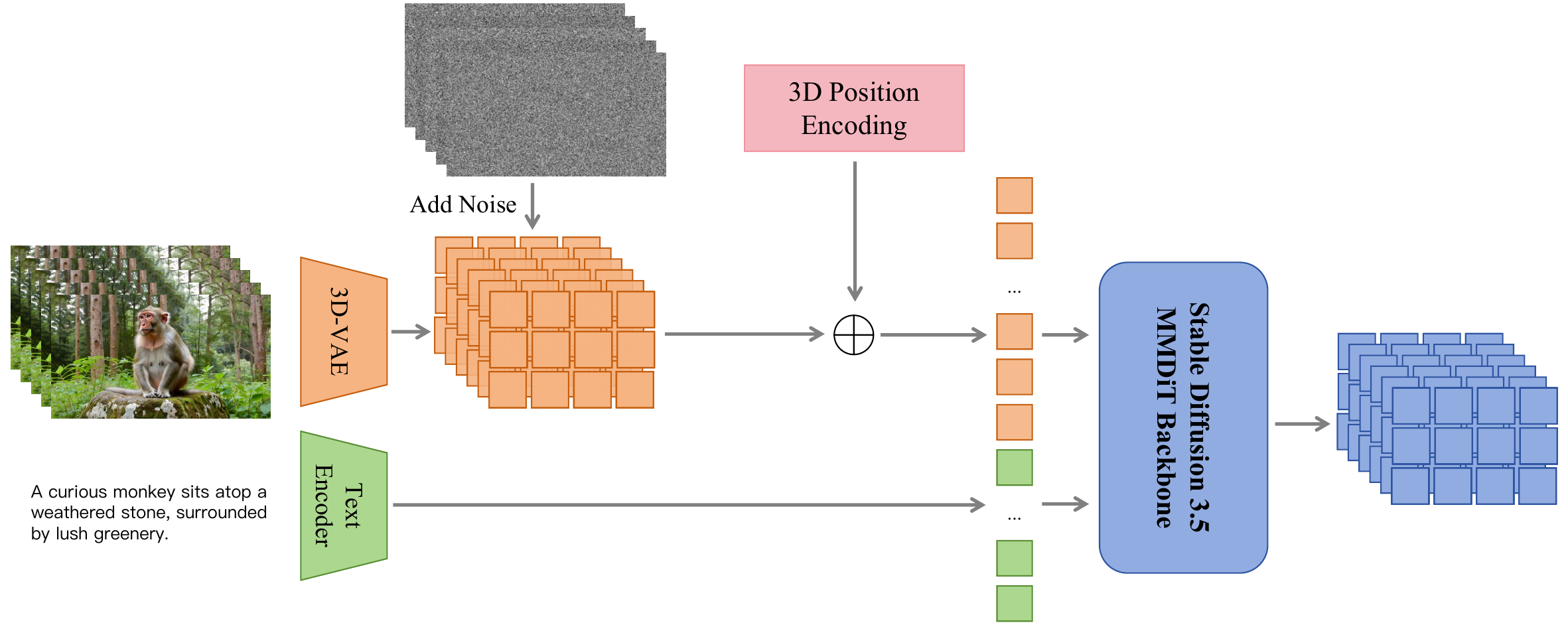}
    \caption{Architecture of \OURMODEL. Simply replace the VAE in SD3.5L with a causal 3D-VAE and incorporate 3D position embedding to unlock the model's video generation capabilities.}
    \label{fig:model_arch}
\end{figure}

As shown in \figrefe{fig:model_arch}, the text-to-video diffusion model based on the SD3.5L~\cite{sd3} architecture also includes three components: variational autoencoder, text encoder, and DiT. The most significant architectural change occurred in the variational autoencoder, where 3D-VAE replaced the 2D-VAE in image generation. This modification enables the model to handle images and videos by compressing them into a unified latent representation. The 3D-VAE first compresses the input visual data, which is then processed through a 3D patchification operation. Subsequently, these patchified features are flattened into a one-dimensional sequence, enabling the DiT model to handle both image and video generation tasks simultaneously. As shown in the~\figrefe{tab:model_config}, the number of attention layers and dimensions of \OURMODEL~are consistent with those of SD3.5L.

\begin{table}[htbp]
\centering
\caption{Architecture configuration of ContentV.}
\label{tab:model_config}
\begin{tabular}{ccccc}
\toprule
\textbf{Patch Size} & \textbf{\#Layers} & \textbf{Attention Heads} & \textbf{Head Dimension} & \textbf{FFN Dimension} \\    
\midrule
1x2x2 & 38 & 38 & 64 & 9728 \\
\bottomrule
\end{tabular}
\end{table}

\subsection{3D Variational Auto-encoder}
Previous works~\cite{lin2024open,yang2024cogvideox,wang2025wan,kong2024hunyuanvideo} use a 3D-VAE to compress pixel-space images and videos into a unified latent space. It is common to adopt temporal causal 3D convolution~\cite{yu2023language}, ensuring the frame only depends on previous frames. Specifically, for a video of shape $(T+1)\times 3 \times H \times W$, the 3D-VAE compresses it into latent space with shape $(\frac{T}{c_{t}}+1)\times C\times(\frac{H}{c_{s}})\times(\frac{W}{c_{s}})$, where $c_{s}$ is spatial downsample factor and $c_{t}$ is temporal downsample factor. Follow previous works, we use $c_t = 4$, $c_s = 8$, and $C = 16$ in our implementation. This compression significantly reduces the number of tokens for long-duration videos, allowing us to train videos at the original resolution and high frame rate. For the open-source version, we directly used the VAE of Wan2.1~\cite{wang2025wan} as the encoder for images and videos.

\subsection{3D Diffusion Transformer}
The DiT of SD3.5L is designed for image input and needs to be modified to adapt to video input. Since the latent feature will be flattened into a one-dimensional sequence, the position information needs to be distinguished through position embedding. Furthermore, for video input, the sequence length becomes significantly longer (\textasciitilde 116k), and QK-Norm plays an important role in stabilization training.

\begin{wrapfigure}{r}{0.35\textwidth}
  \centering
  \includegraphics[width=0.35\textwidth]{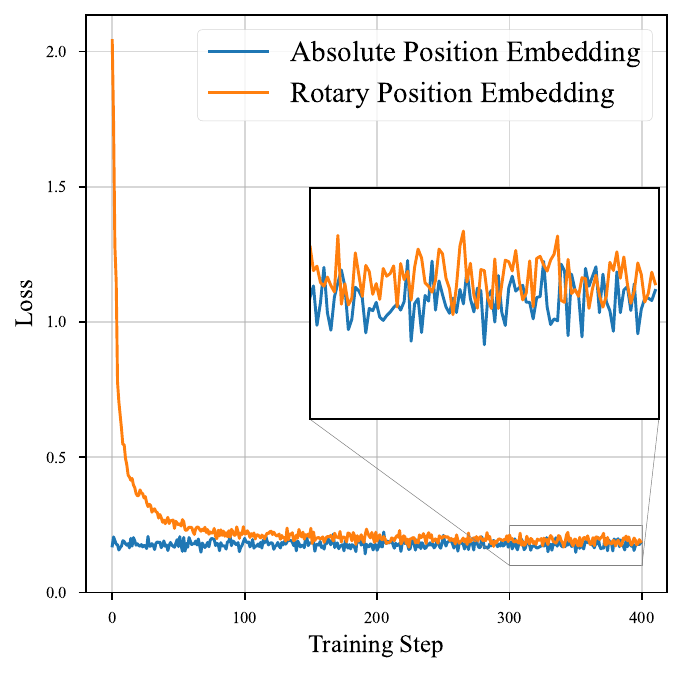}
  \vspace{-10pt}
  \caption{RoPE v.s. APE}
  \label{fig:rope_abe_loss}
  \vspace{-10pt}
\end{wrapfigure}
\paragraph{3D Position Embedding} In SD3.5L~\cite{sd3}, \textit{Absolute Position Embedding} (APE) is used to capture spatial relationships. When extending the DiT of SD3.5L from image input to video input, we add additional time position embeddings to appropriately represent sequential time information. Recently, \textit{Rotary Position Embedding} (RoPE) uses rotation matrix for position encoding, which enables the relative position information between tokens at different positions to be well learned, improves the model's adaptability to long context, and has become the mainstream method for large language model and text-to-video diffusion models. 
We also attempted to replace APE in SD3.5L with RoPE. The loss curve is shown in the ~\figrefe{fig:rope_abe_loss}. The model only needs to be trained with a small number of steps (\textasciitilde 500) to quickly adapt to the new position encoding. And after longer training, RoPE did not show a significant improvement in its score on VBench compared to APE. Considering the marginal gap between RoPE and APE, we still retained APE in text-to-video setting.

\paragraph{Q-K Normalization} In large language model, training large-scale transformers can occasionally result in loss spikes, which may lead to model corruption. 
Training DiT on long sequences also faces the same challenges, especially when training with mixed precision. Half-precision can lead to gradient explosion in long sequences, and the growth of grad norm causes the loss to not converge. Single-precision training using float32 can avoid this problem, but the training time increases significantly. The query-key normalization~\cite{dehghani2023scaling} used in SD3.5L can avoid this problem. Specifically, we apply RMSNorm~\cite{zhang2019root} to each query-key feature prior to attention computation, ensuring smoother and more reliable training dynamics. 

\section{Training Strategy}

\subsection{Flow Matching}
\paragraph{Training Objective} The \OURMODEL~is trained using Flow Matching~\cite{lipman2022flow}, an algorithm that enables efficient sampling through straight probability paths in continuous time. 
Given paired data samples $\bm{x}_1 \sim p_{\text{data}}$ and noise samples $\bm{x}_0 \sim \mathcal{N}(\mathbf{0},\mathbf{I})$, we define the interpolated sample at time $t \in [0,1]$ as $\bm{x}_t = (1-t)\bm{x}_0 + t\bm{x}_1$. The model is trained to predict the velocity $v_\text{target}=\mathrm{d}\bm{x}_t/\mathrm{d} t=\bm{x}_1-\bm{x}_0$, which guides the $\bm{x}_t$ towards $\bm{x}_1$. The model parameters are optimized by minimizing the mean squared error between the predicted velocity $\bm{v}_\theta$ and the ground truth velocity $\bm{v}_\text{target}$, expressed as the loss function:
\begin{equation}
\mathcal{L}_{\text{FM}} = \mathbb{E}_{t,\bm{x}_0,\bm{x}_1} \left[ \| \bm{v}_\theta(\bm{x}_t, t) - \bm{v}_{\text{target}}(\bm{x}_t, \bm{x}_0, \bm{x}_1) \|_2^2 \right]
\label{eq:fm}
\end{equation}
During the sampling, a gaussian noise $\bm{x}_0 \sim \mathcal{N}(\mathbf{0},\mathbf{I})$ will be drawn at first. The first-order euler ordinary differential equation solver is used to compute $\bm{x}_1$ by the velocity prediction. This process ultimately generates the final sample $\bm{x}_1$.

\begin{figure}[htbp]
  \centering
  \begin{subfigure}[b]{0.4\textwidth}
    \includegraphics[width=\textwidth]{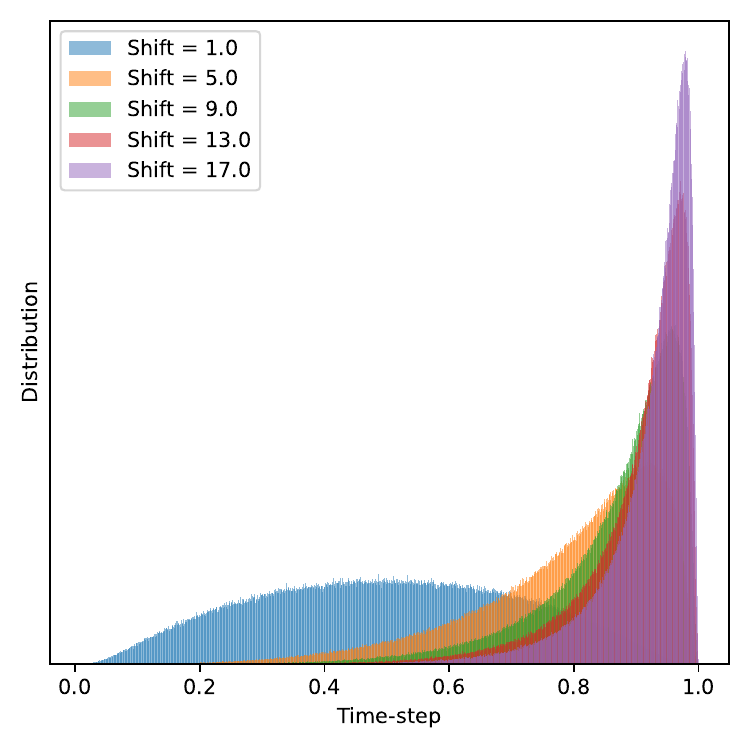}
    \caption{Distribution of timesteps with various shift.}
  \end{subfigure}
  \hfill
  \begin{subfigure}[b]{0.4\textwidth}
    \includegraphics[width=\textwidth]{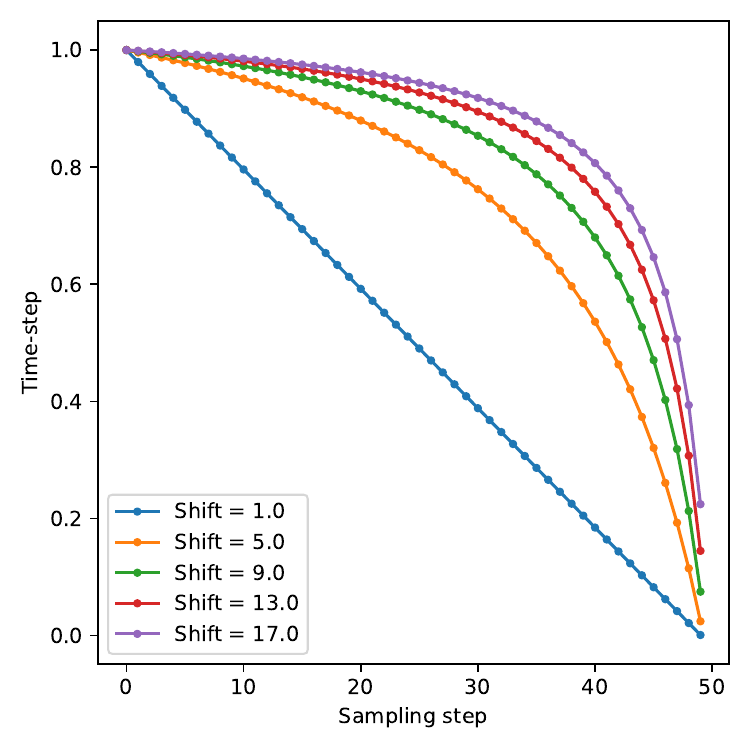}
    \caption{Sampling timestep under varying shift.}
  \end{subfigure}
  \caption{Illustration of Flow shift.}
  \label{fig:shift}
\end{figure}

\begin{figure}[htbp]
  \centering
  \caption{Sampling results under varying classifier-free guidance scales and flow shift.}
  \label{fig:flow_shift_sample}
  \footnotesize
\begin{tabular}{
    >{\raggedright\arraybackslash}m{1.3cm}
    *{5}{>{\centering\arraybackslash}m{2cm}}
    }
  \toprule
     & \textbf{Shift = 1} & \textbf{Shift = 5} & \textbf{Shift = 9} & \textbf{Shift = 13} & \textbf{Shift = 17} \\
  \midrule

    \textbf{CFG = 3} &
    \includegraphics[width=0.15\textwidth]{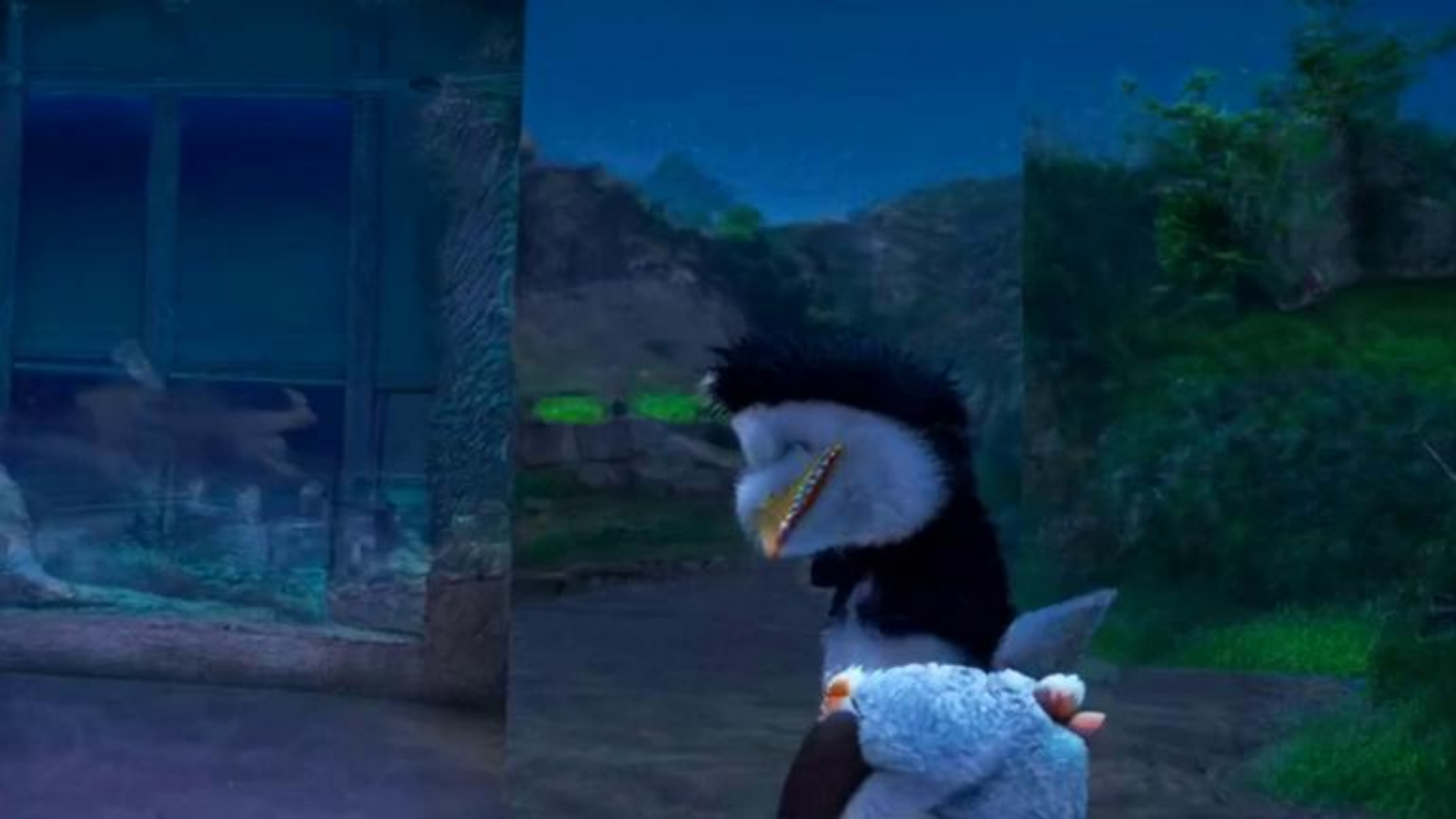} &
    \includegraphics[width=0.15\textwidth]{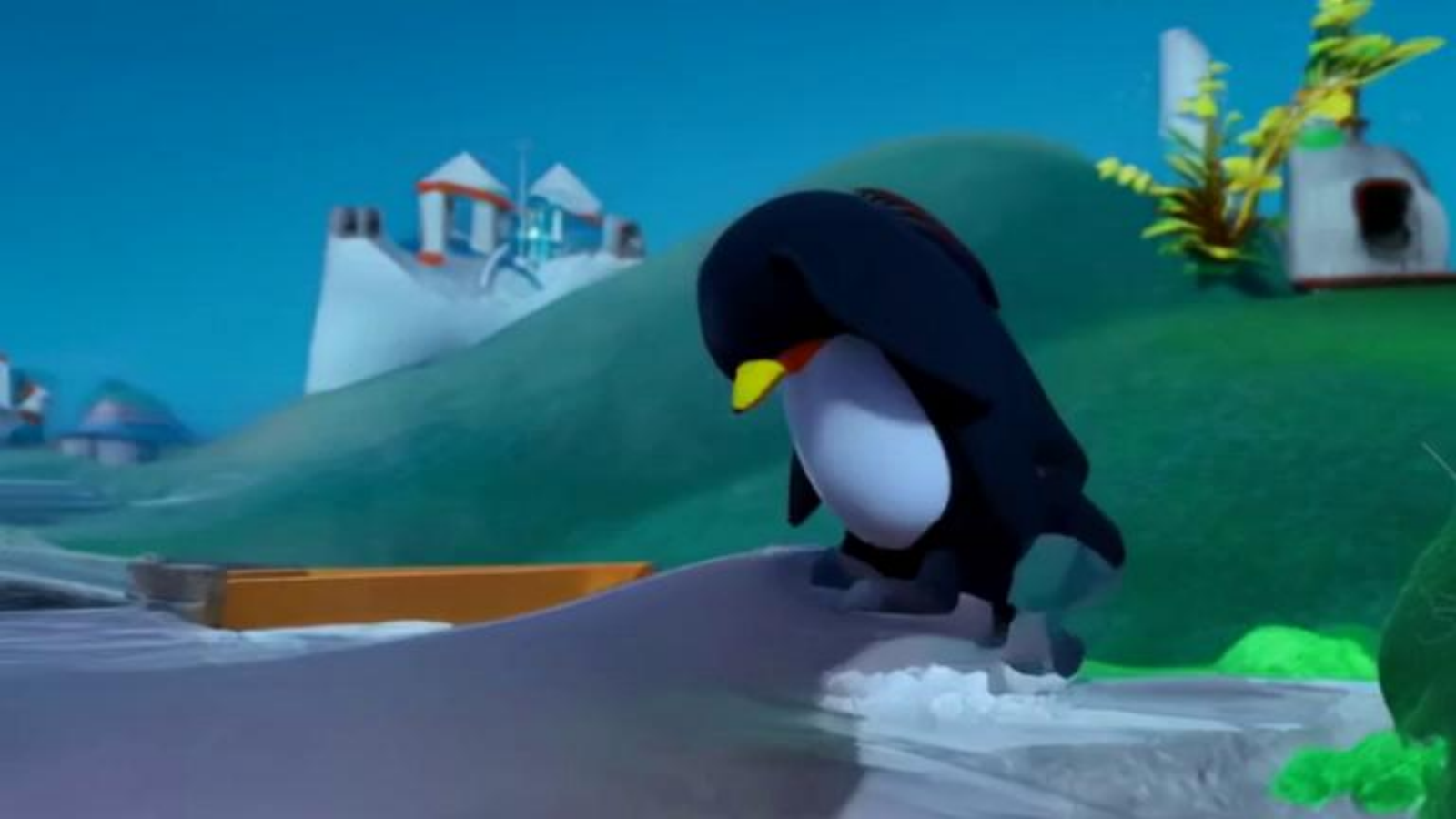} &
    \includegraphics[width=0.15\textwidth]{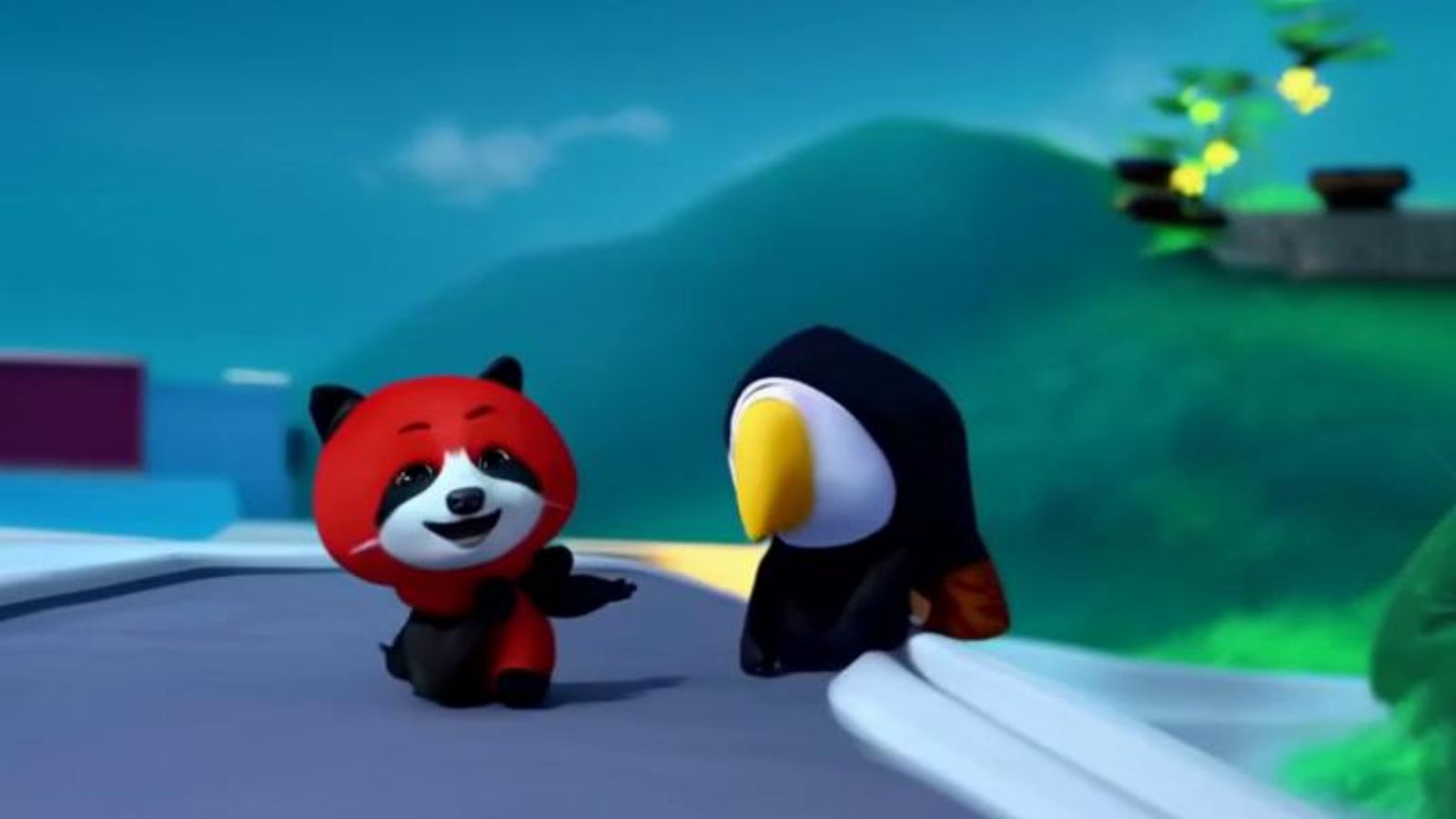} &
    \includegraphics[width=0.15\textwidth]{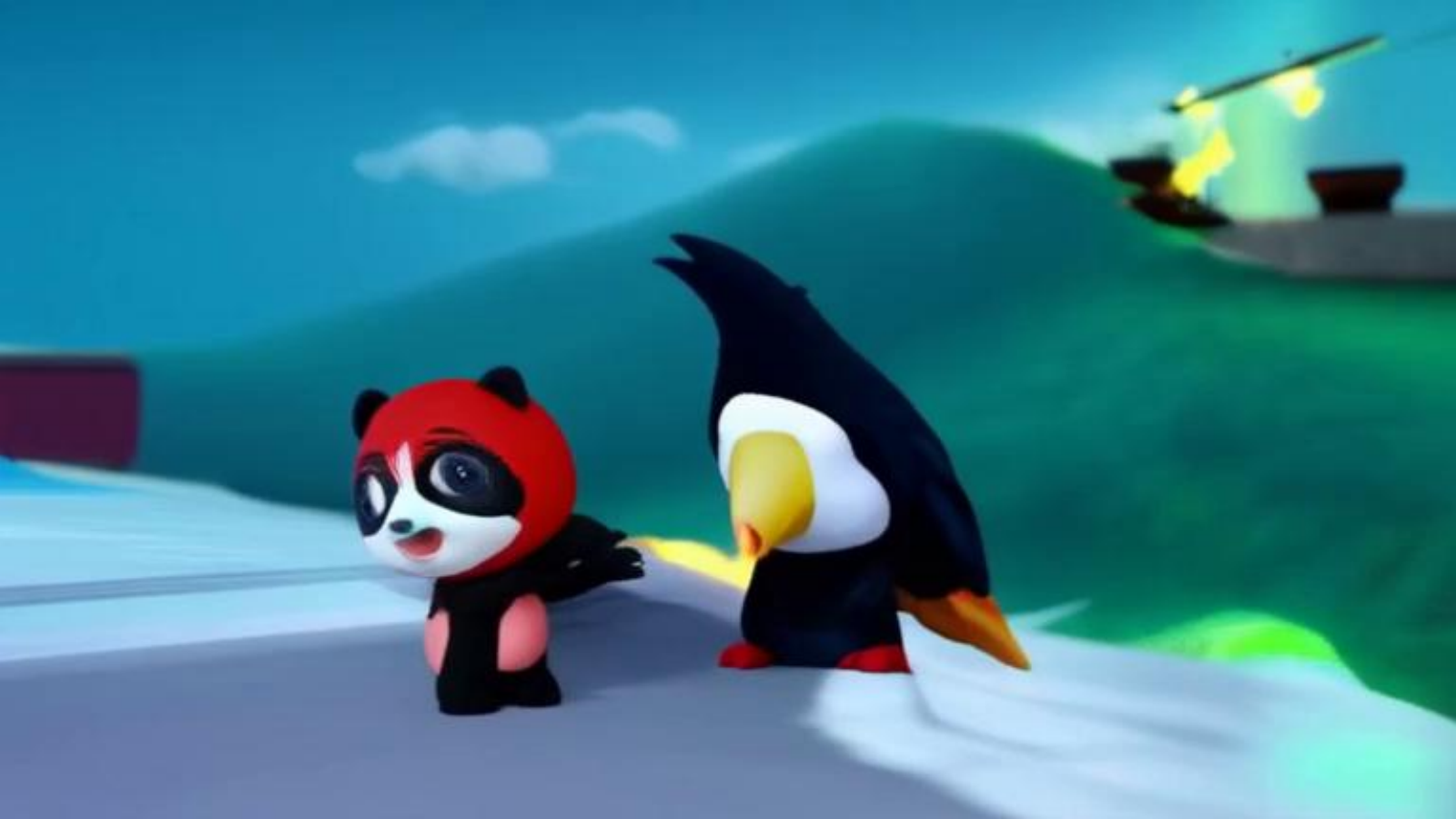} &
    \includegraphics[width=0.15\textwidth]{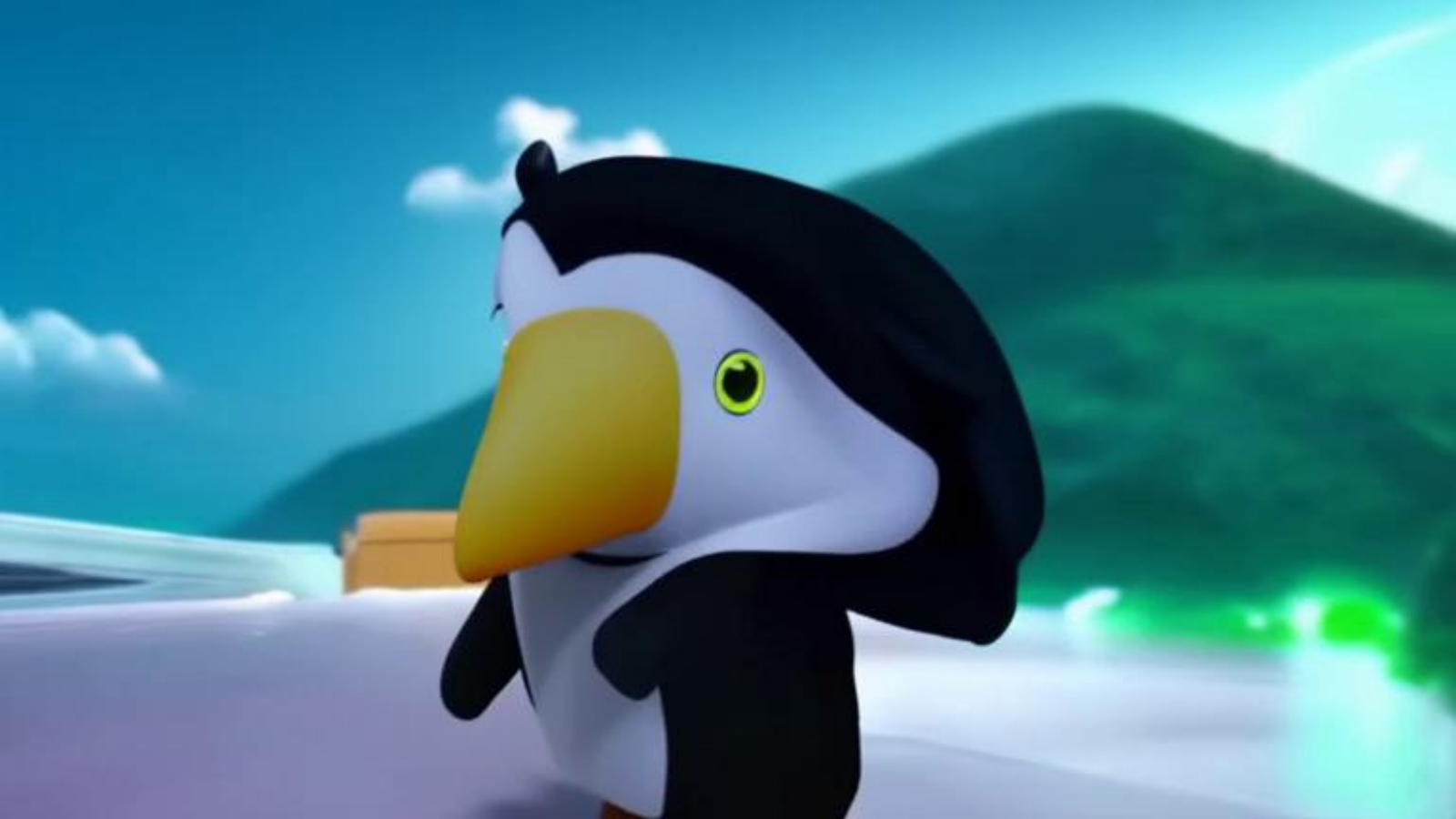} \\
    
    \textbf{CFG = 6} &
    \includegraphics[width=0.15\textwidth]{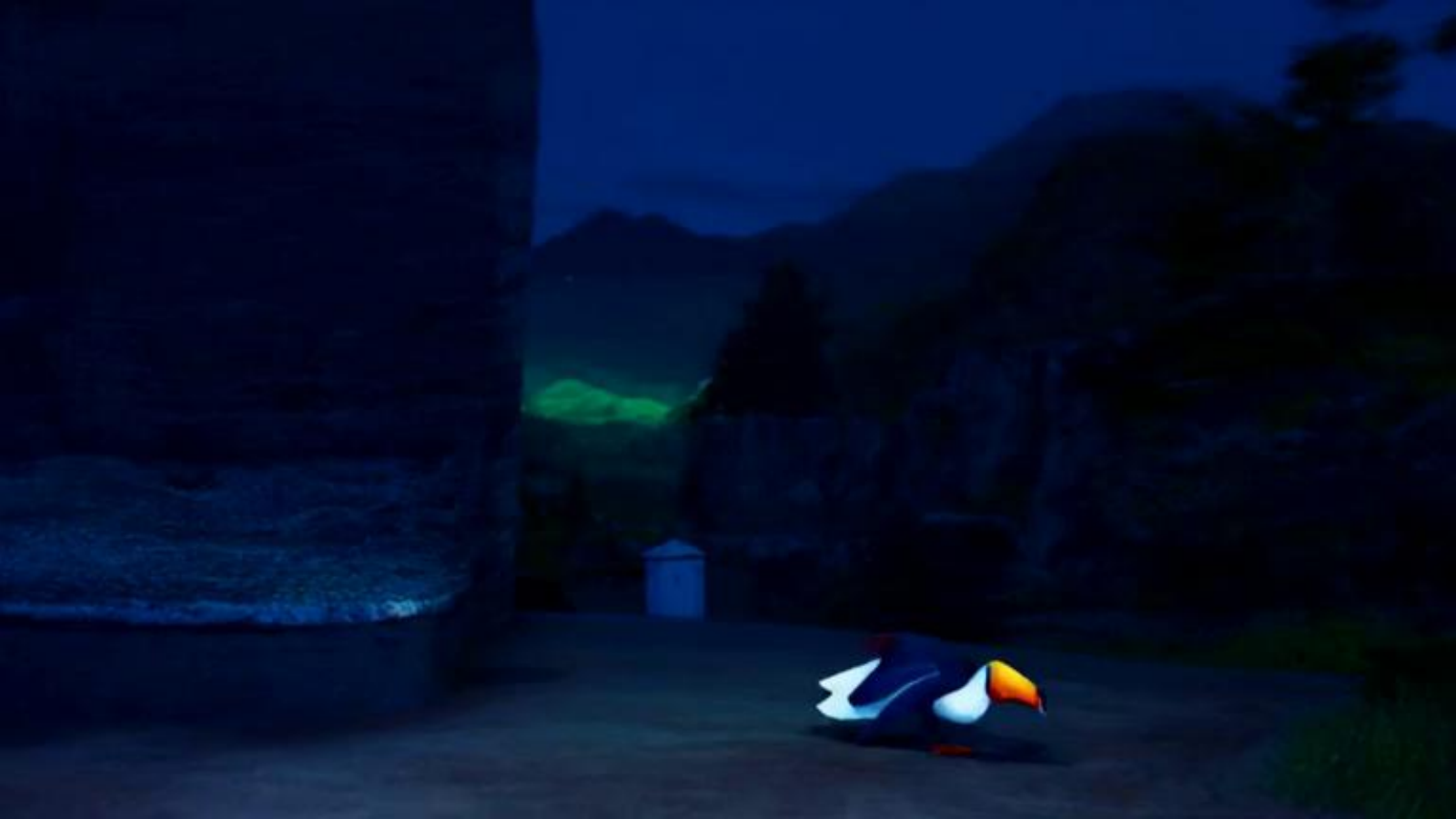} &
    \includegraphics[width=0.15\textwidth]{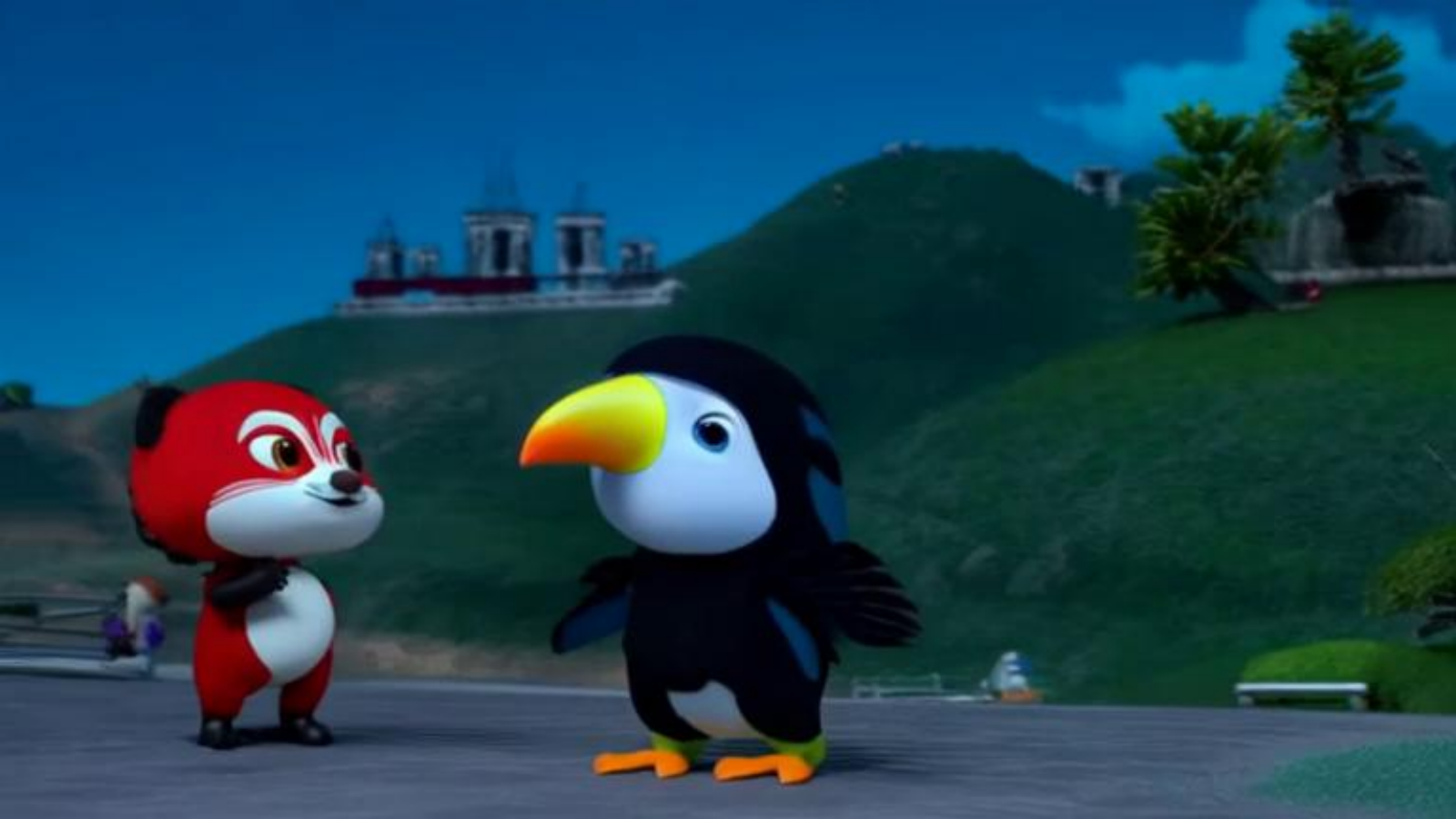} &
    \includegraphics[width=0.15\textwidth]{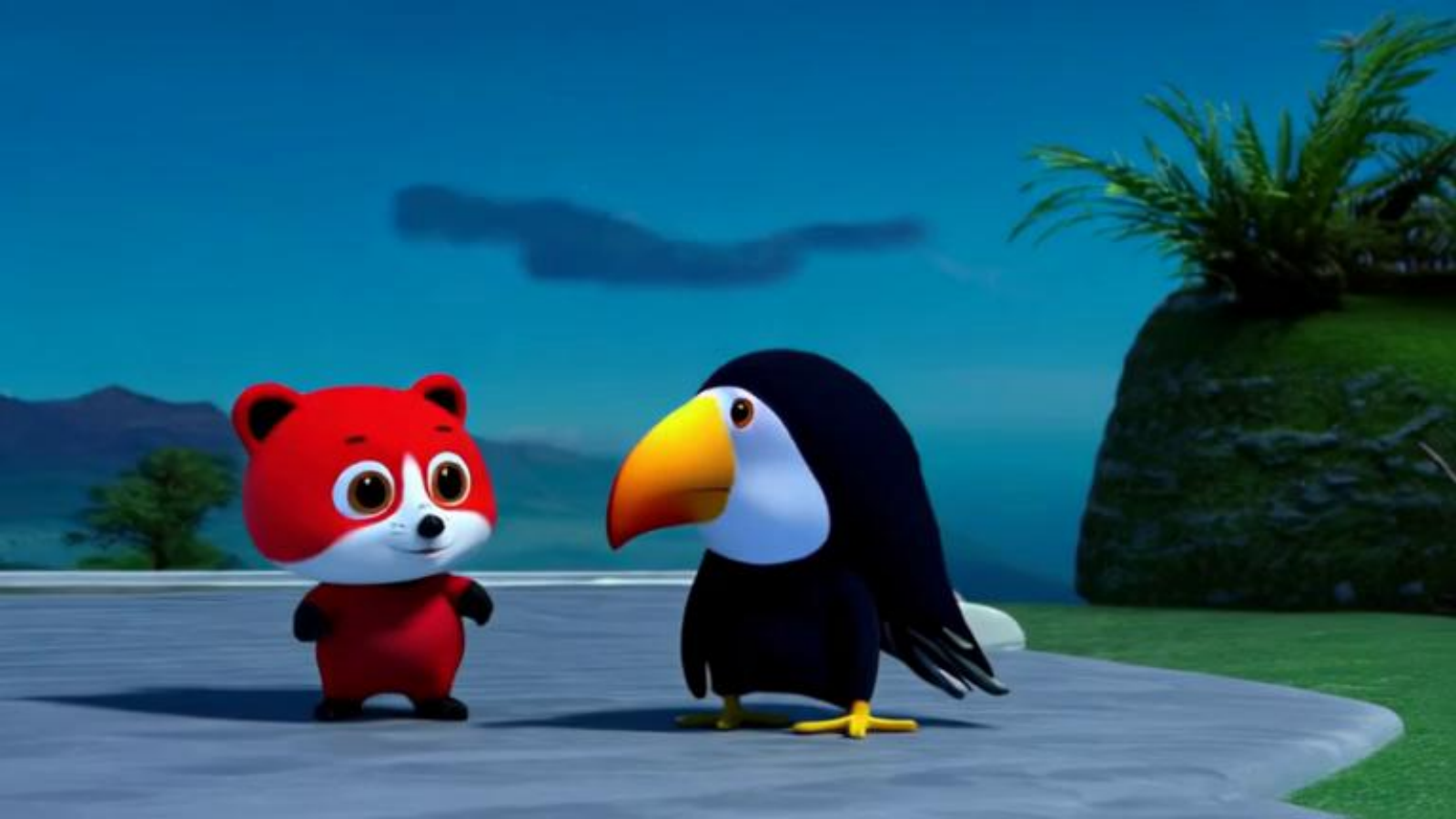} &
    \includegraphics[width=0.15\textwidth]{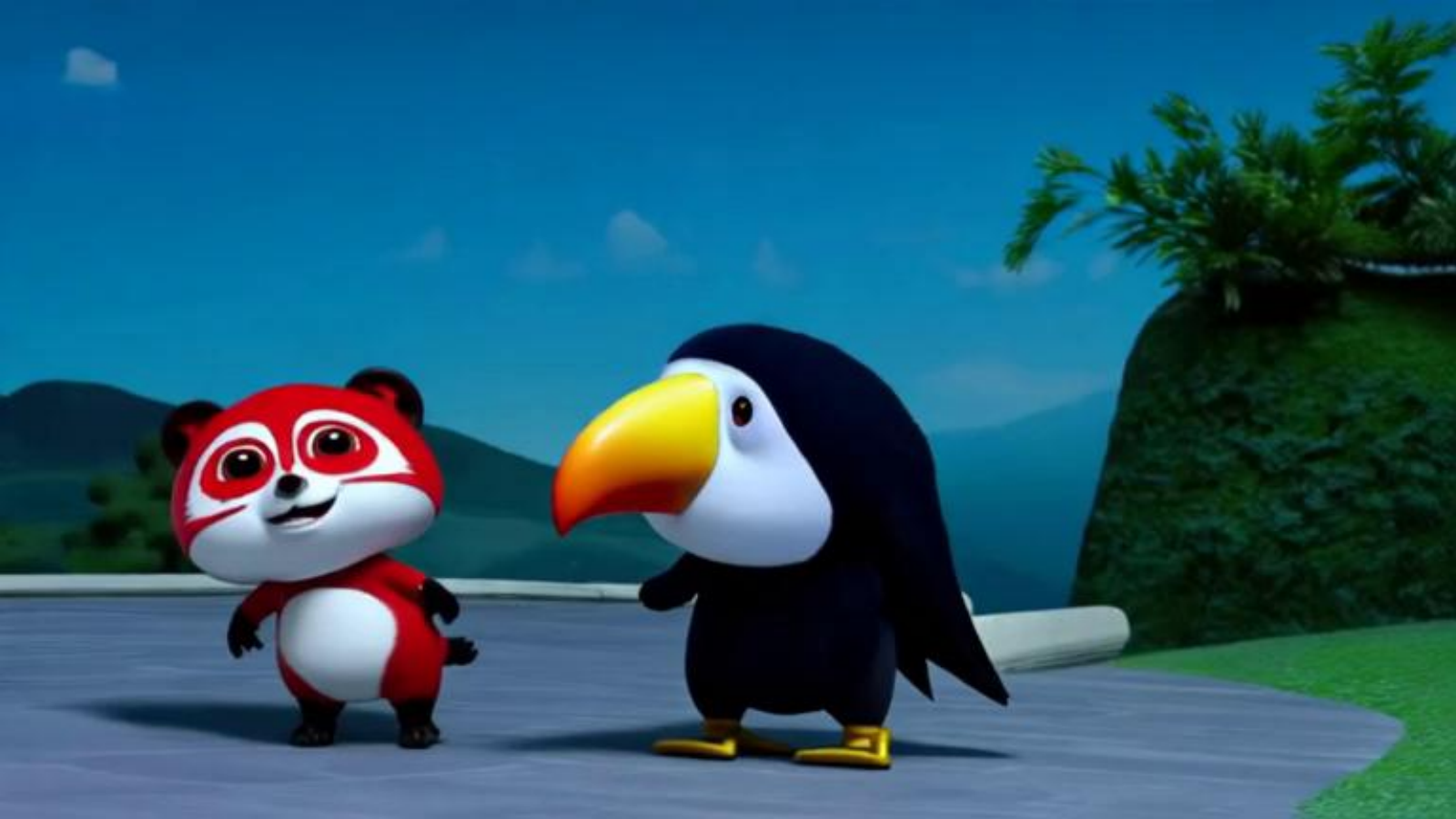} &
    \includegraphics[width=0.15\textwidth]{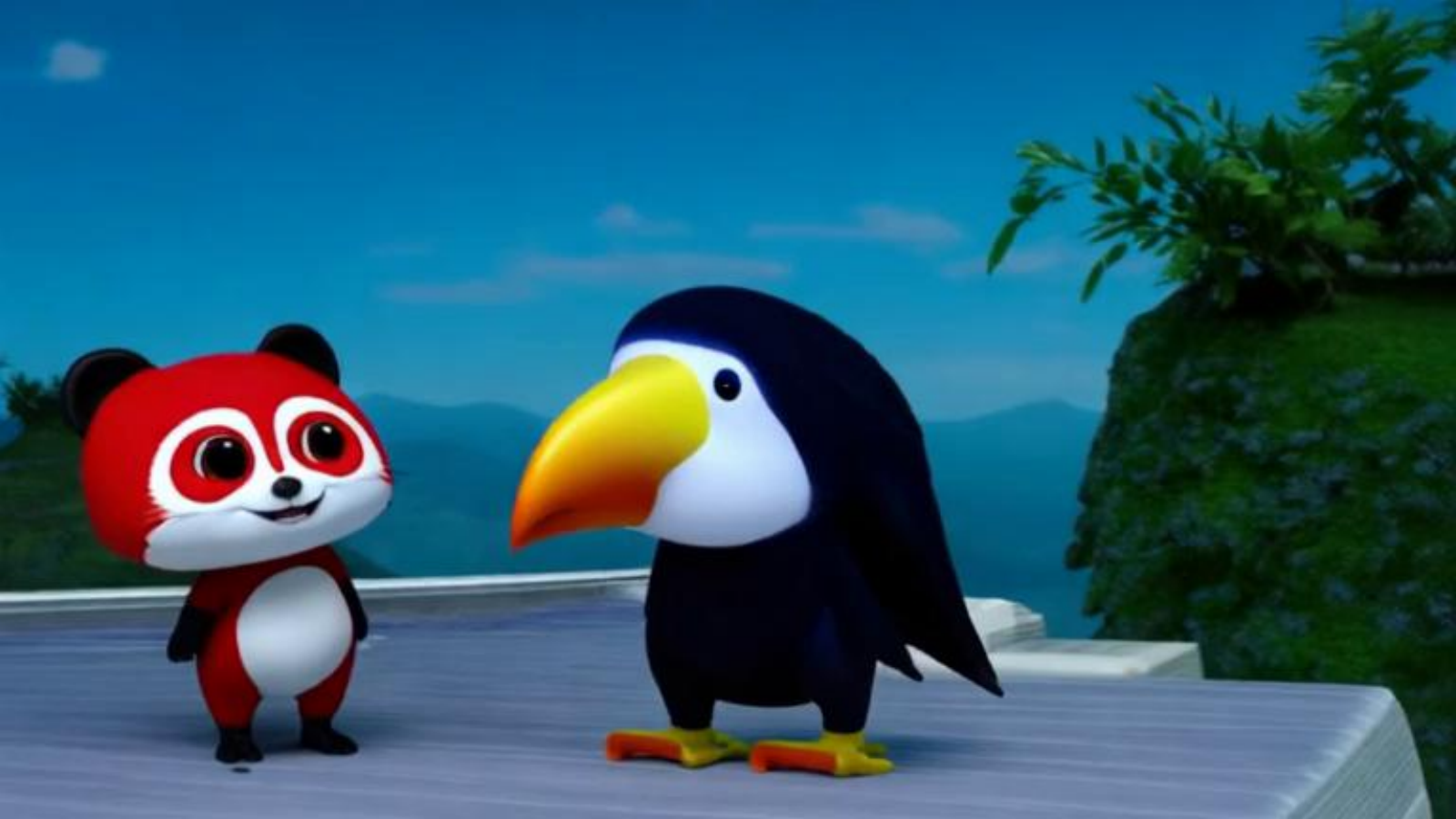} \\
    
    \textbf{CFG = 9} &
    \includegraphics[width=0.15\textwidth]{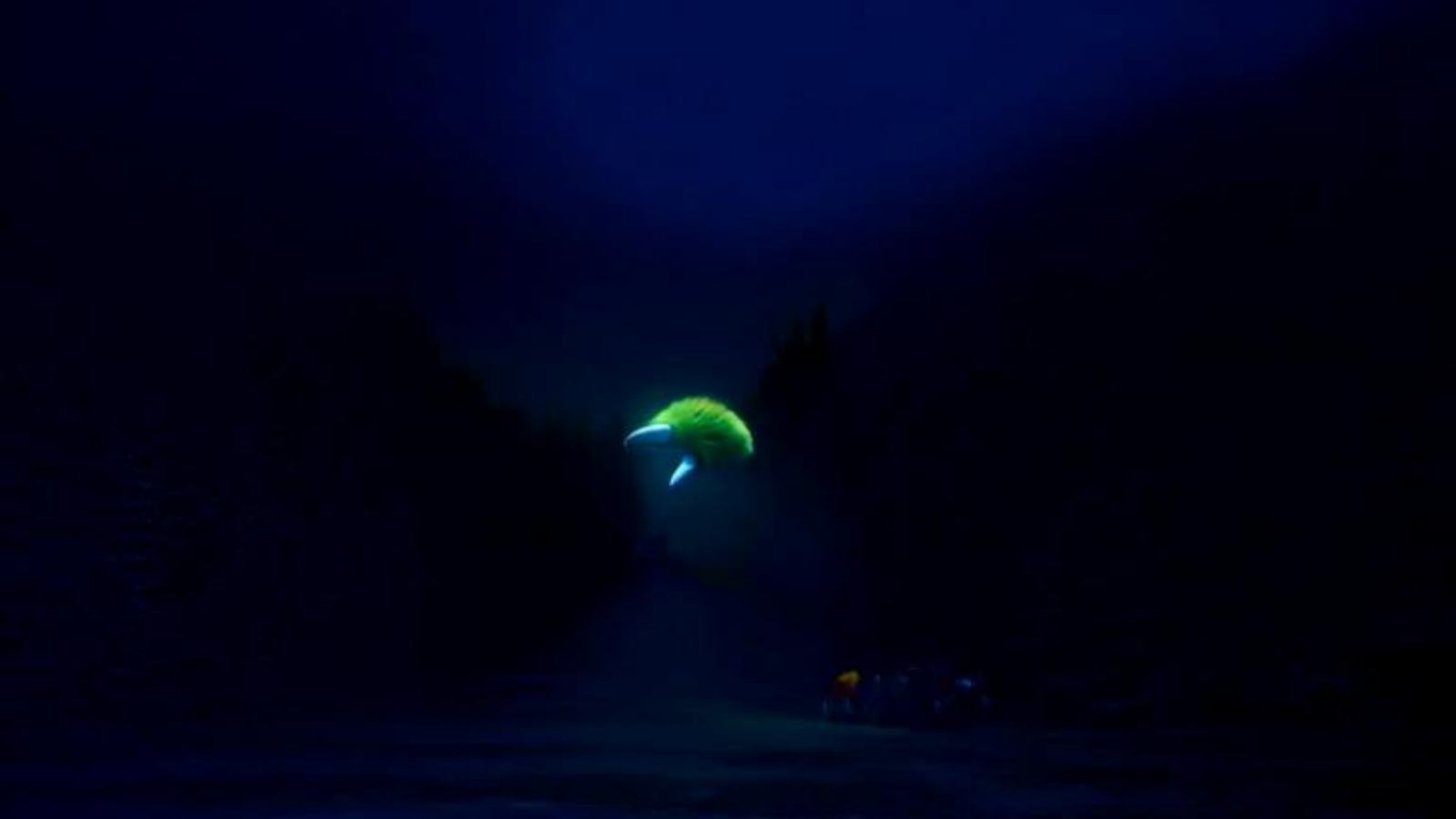} &
    \includegraphics[width=0.15\textwidth]{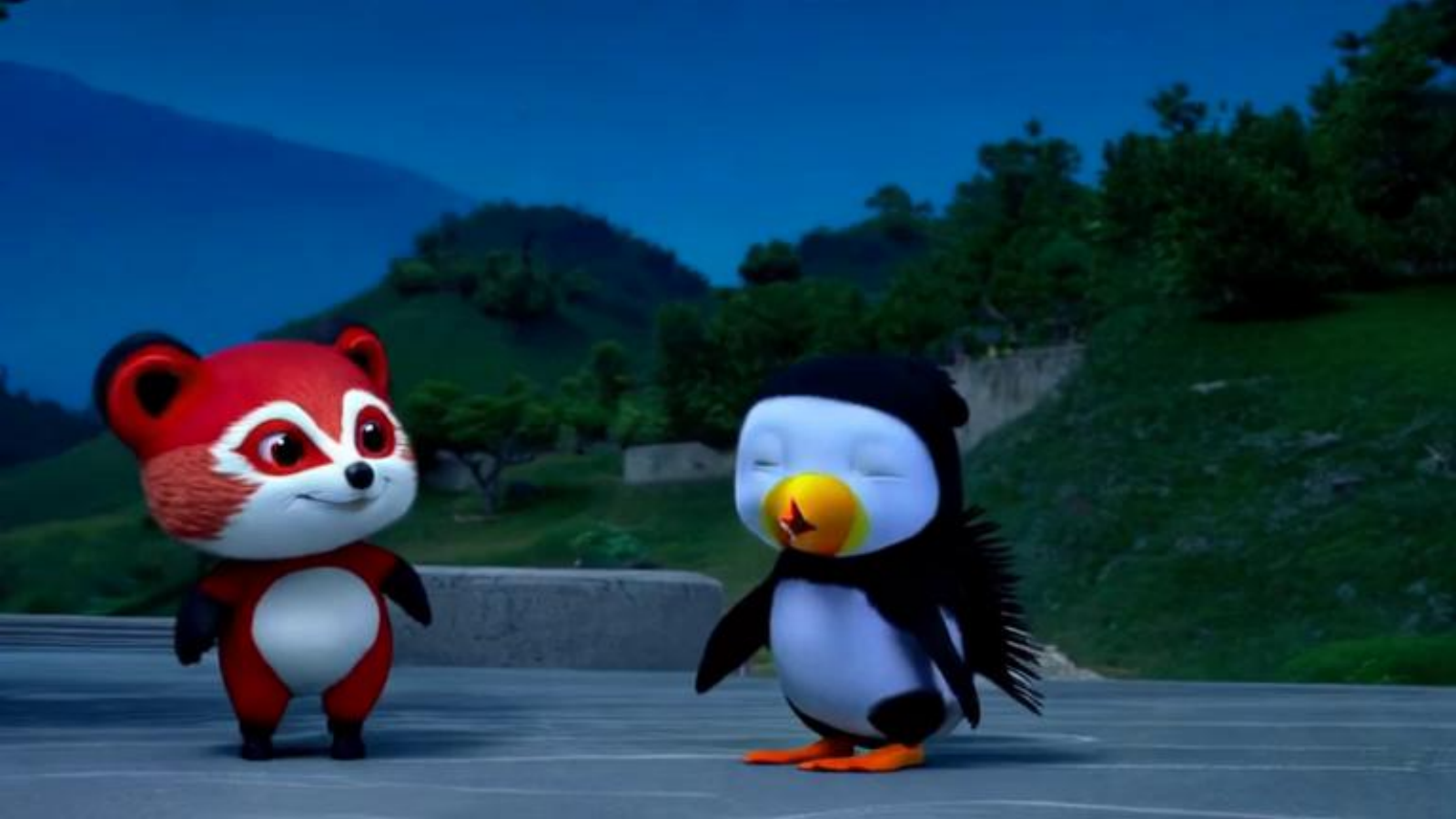} &
    \includegraphics[width=0.15\textwidth]{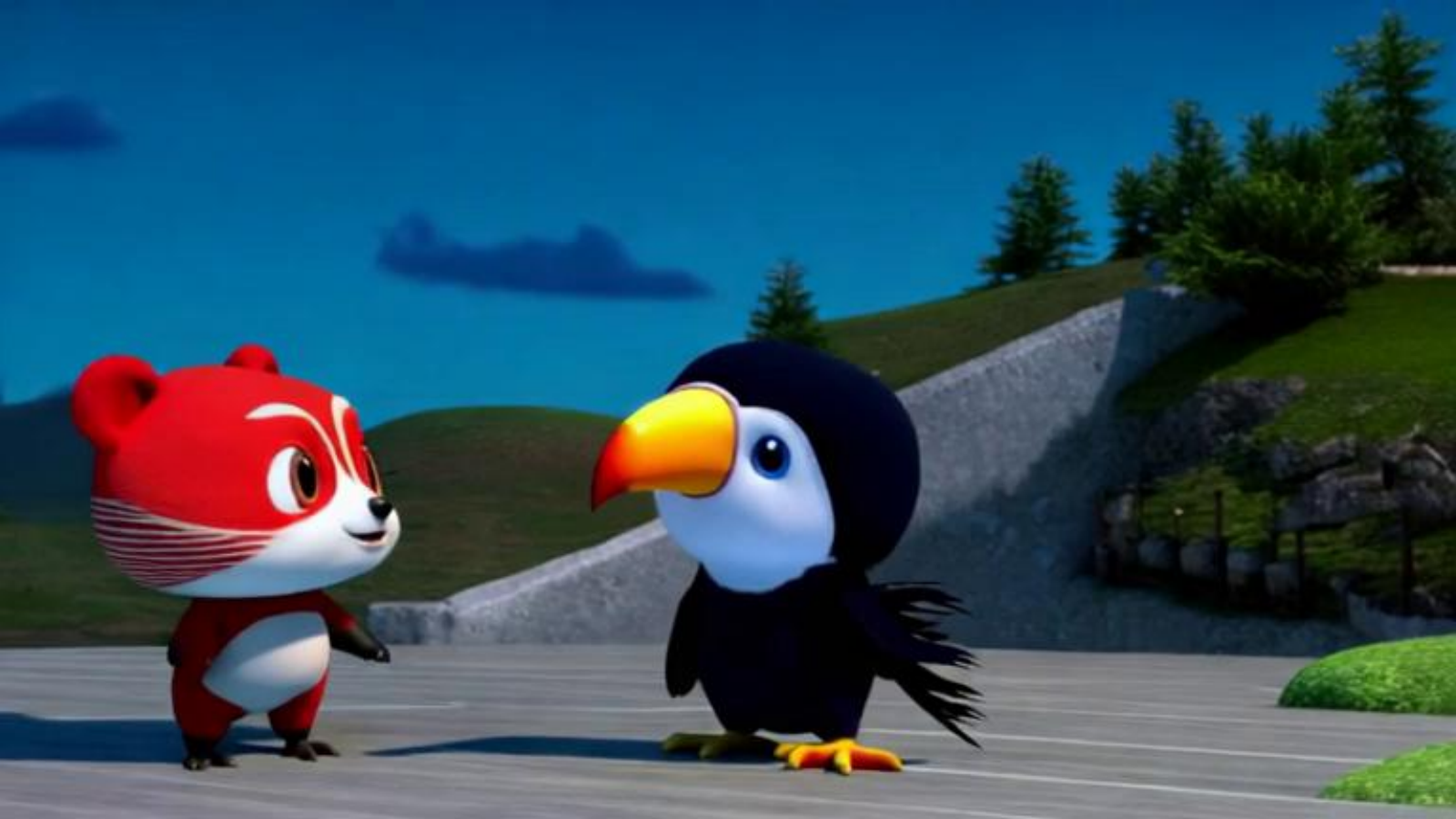} &
    \includegraphics[width=0.15\textwidth]{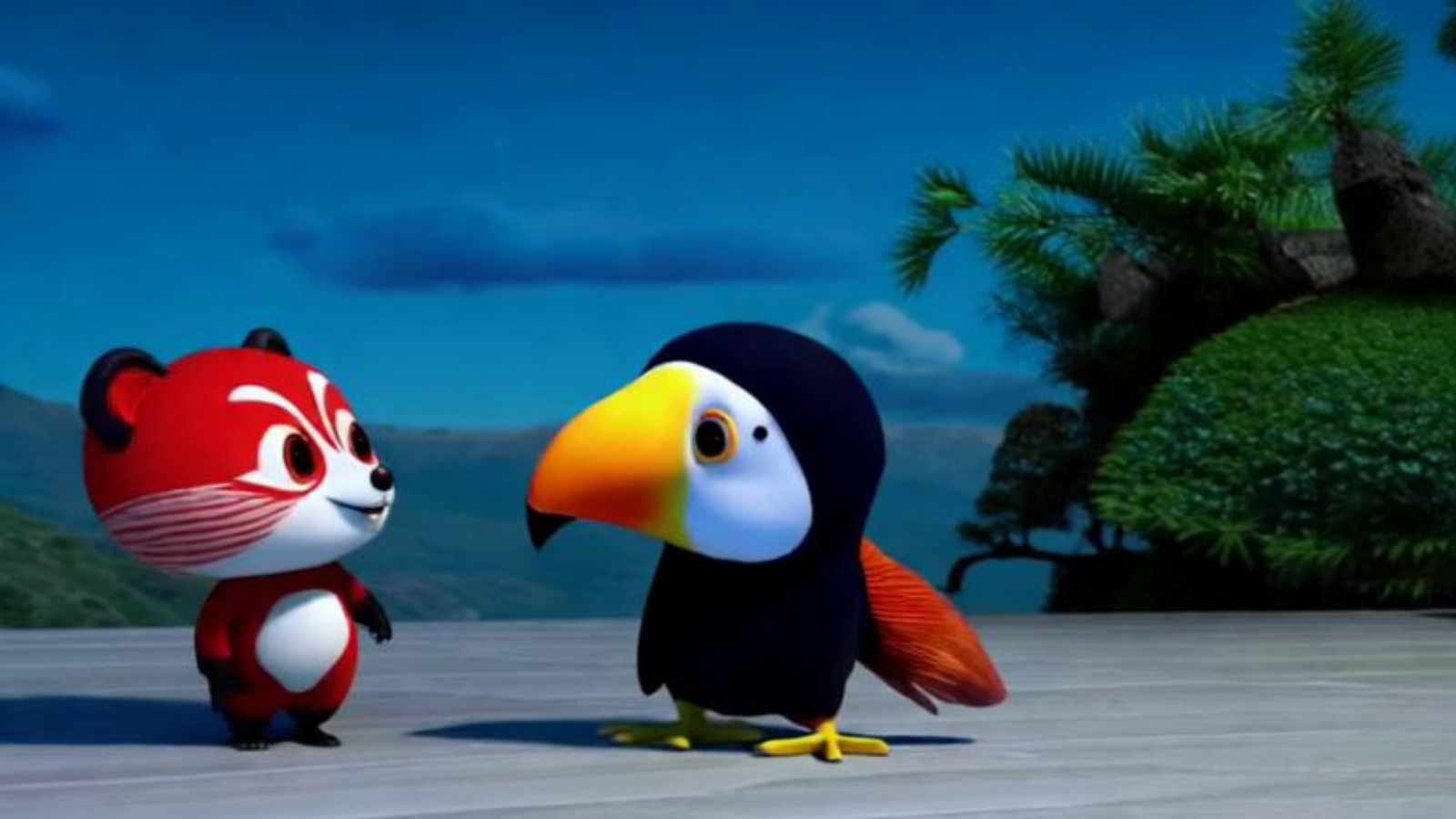} &
    \includegraphics[width=0.15\textwidth]{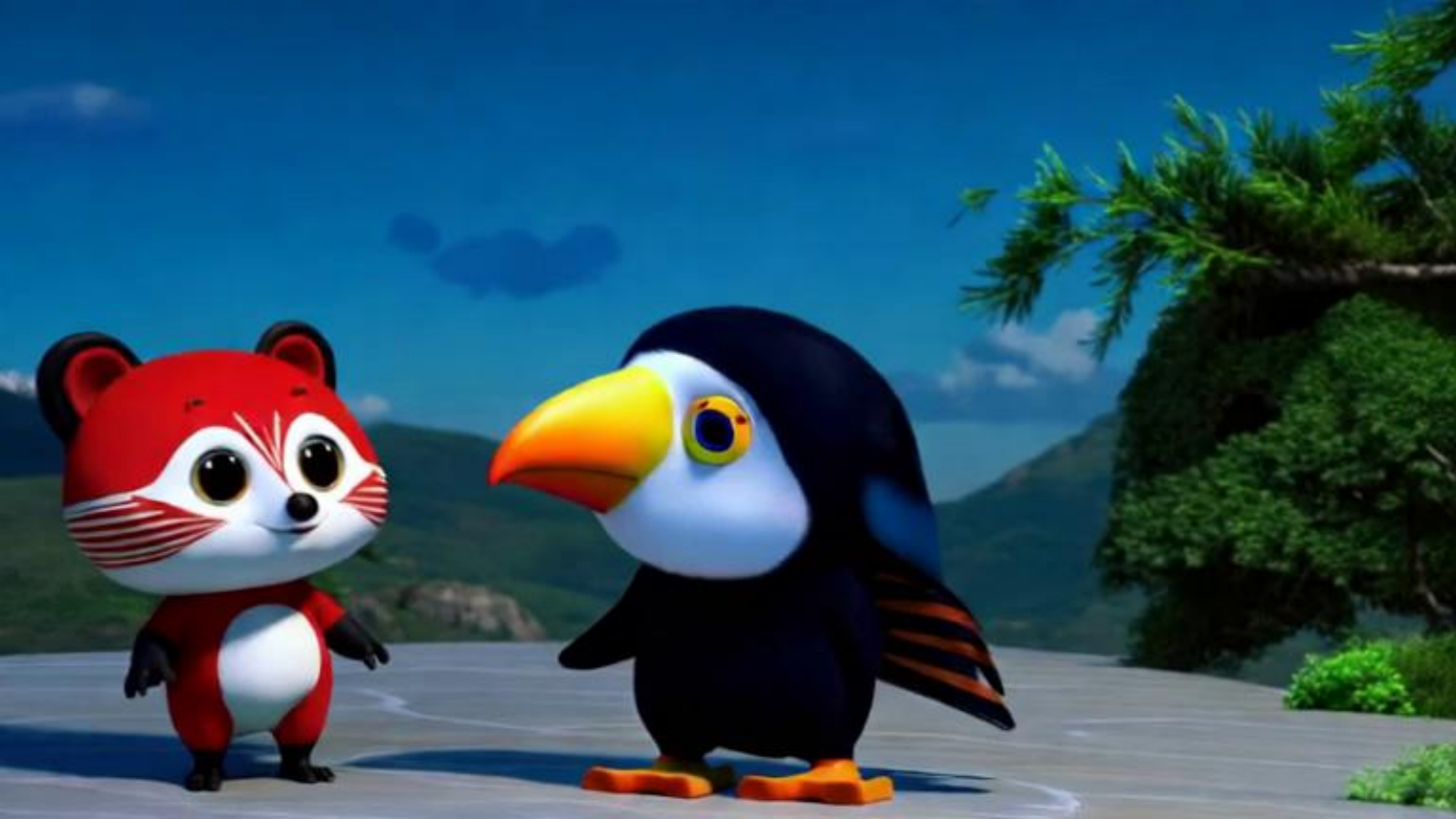} \\
    
    \textbf{CFG = 12} &
    \includegraphics[width=0.15\textwidth]{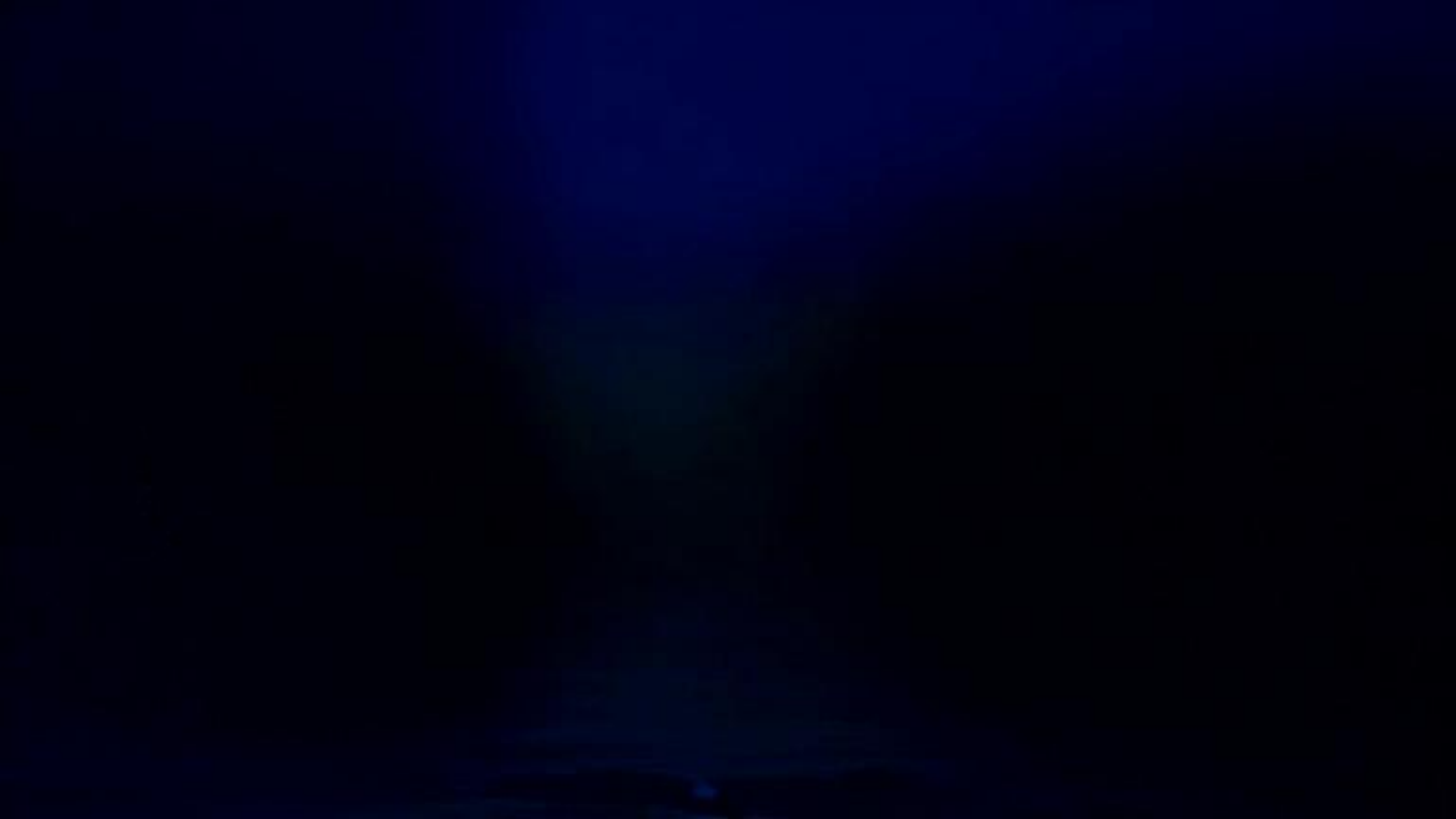} &
    \includegraphics[width=0.15\textwidth]{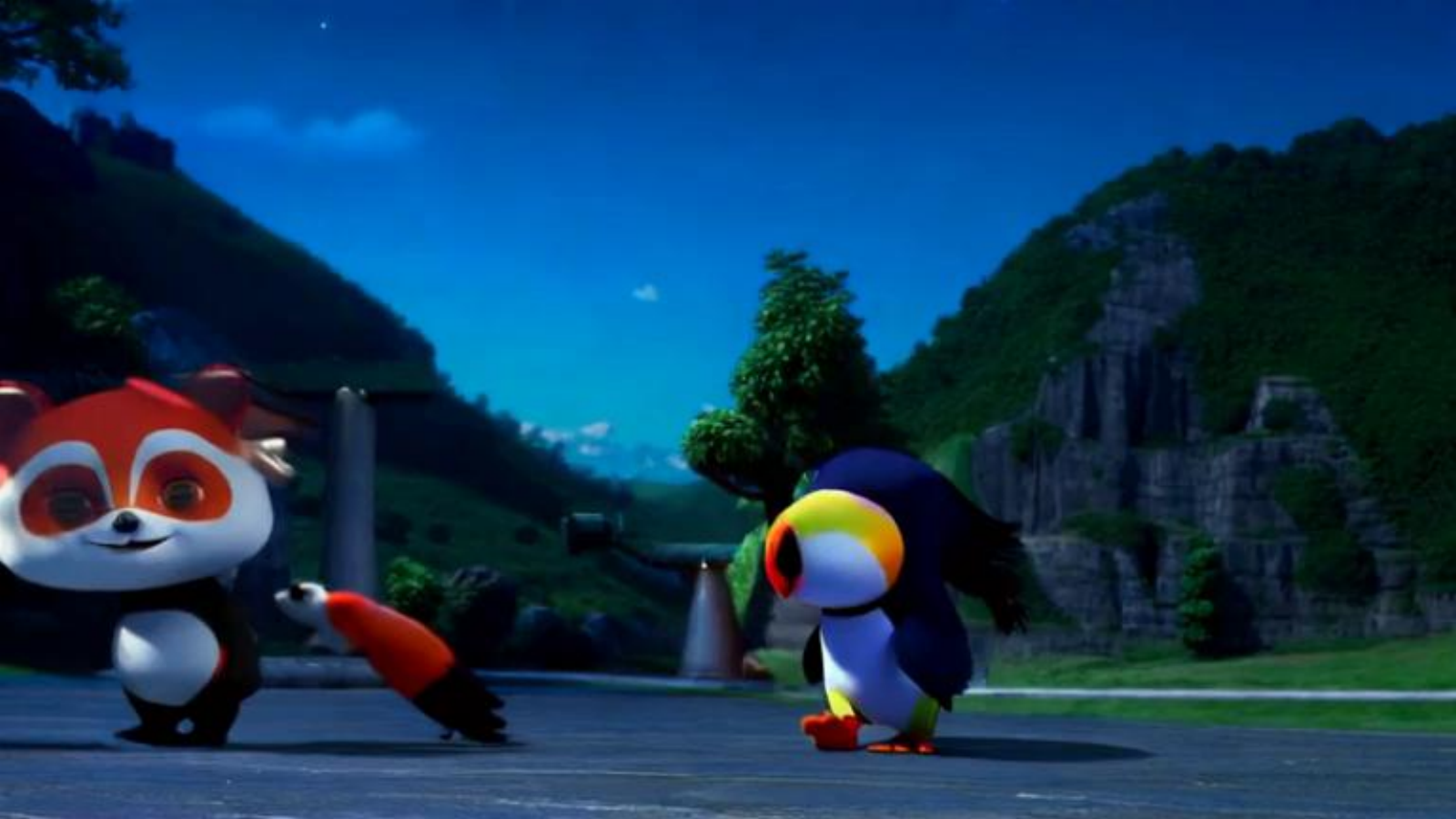} &
    \includegraphics[width=0.15\textwidth]{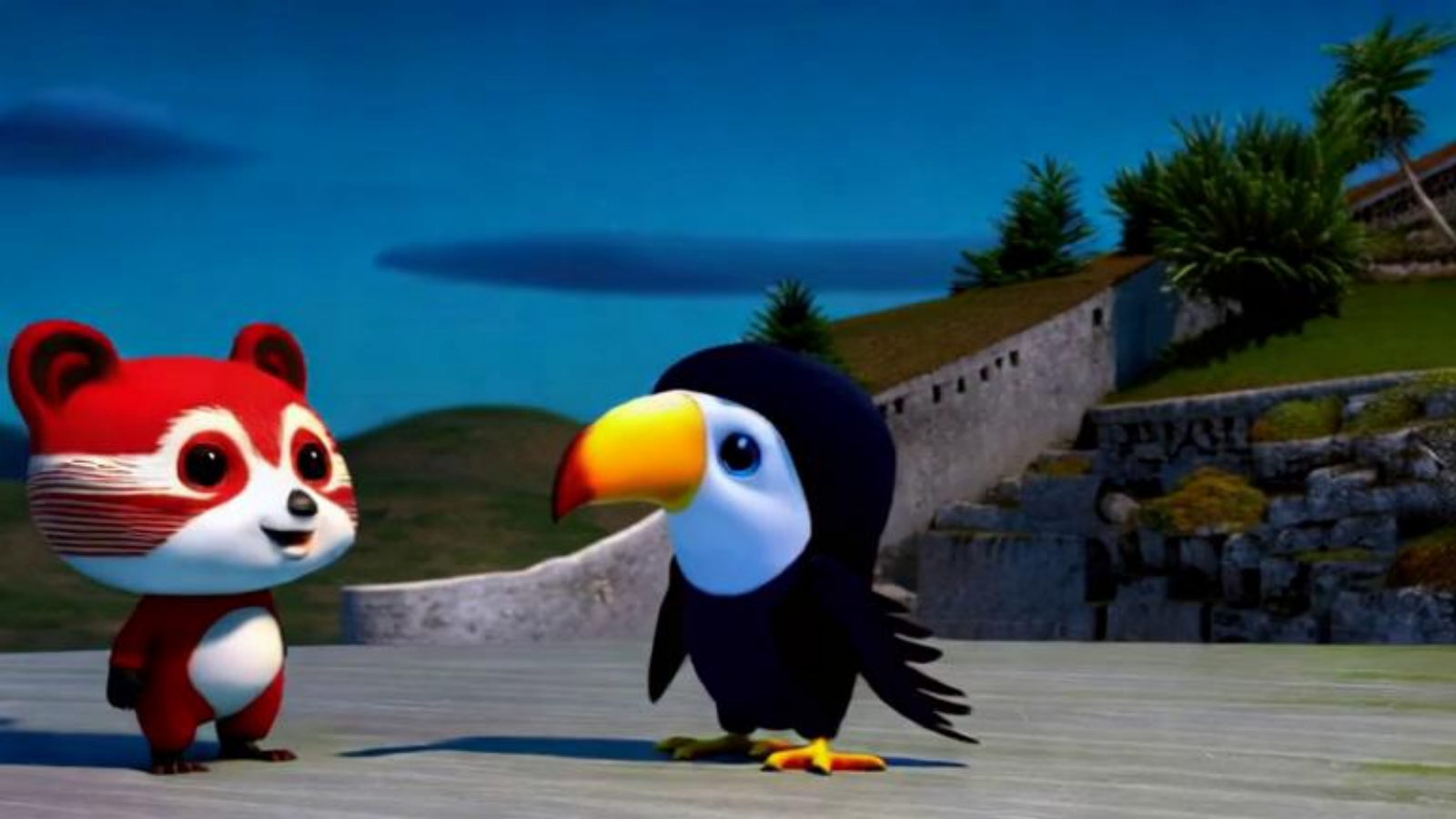} &
    \includegraphics[width=0.15\textwidth]{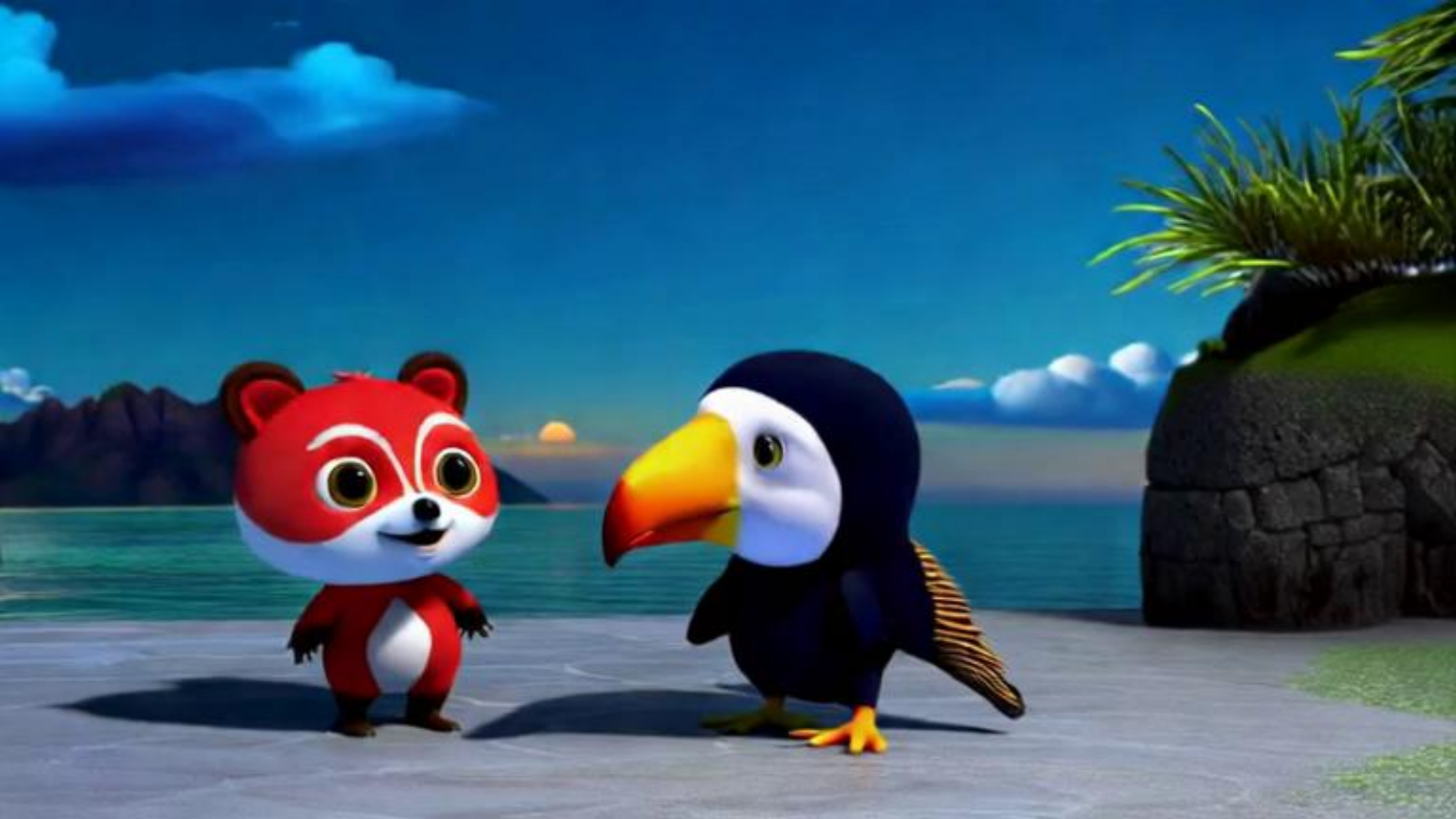} &
    \includegraphics[width=0.15\textwidth]{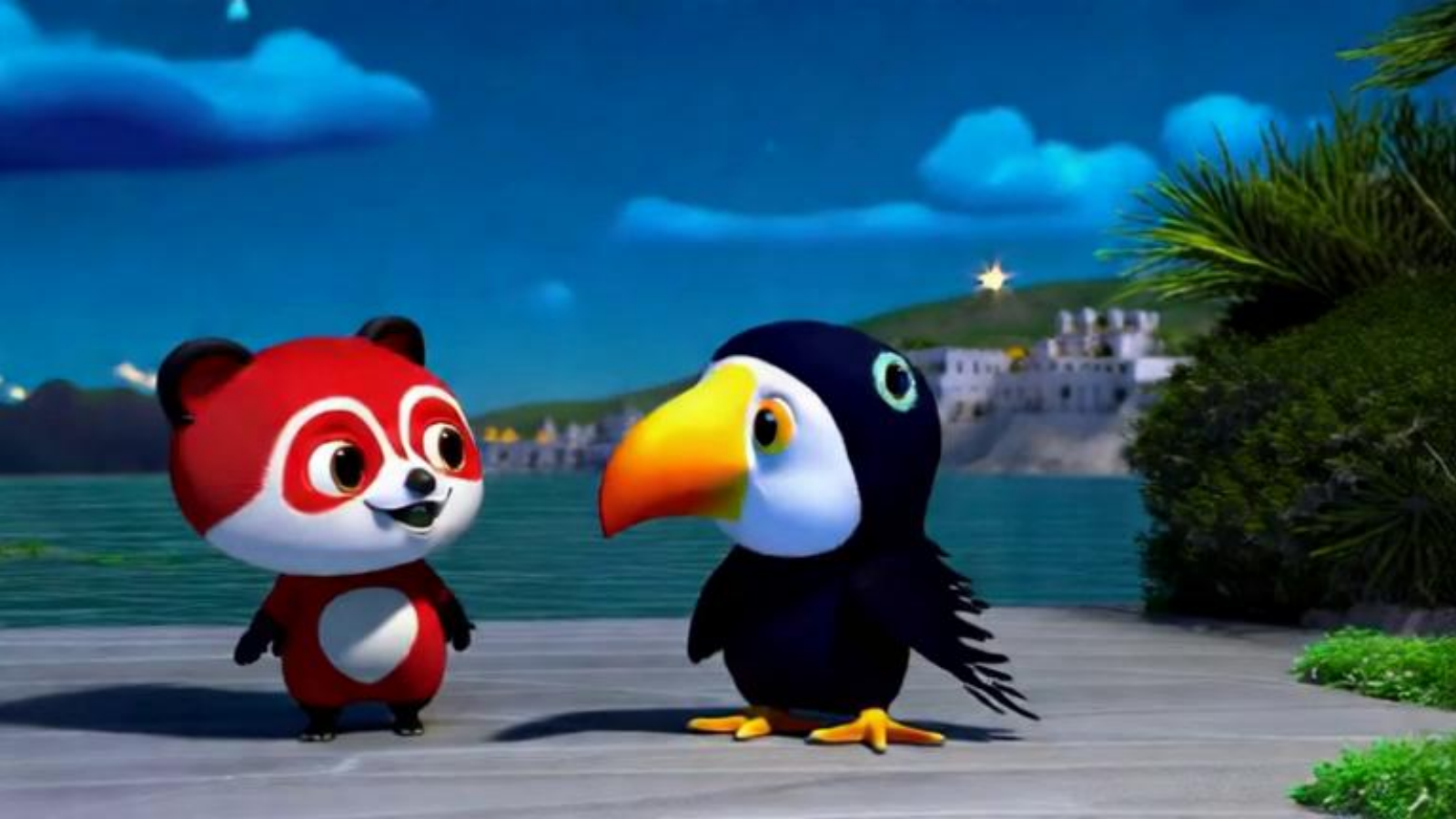} \\
  \bottomrule
  \end{tabular}

\end{figure}

\paragraph{Flow Shift} The $t$ in \eqrefe{eq:fm} is initially sampled from uniform distribution. However, \citet{sd3} demonstrated that directly applying the same noise schedule from low-resolution to high-resolution image generation leads to insufficient noise addition and inadequate image degradation. They proposed using logit-normal and flow shift to replace the uniform distribution. As shown in ~\figrefe{fig:shift}, during training, this timestep sampling method is equivalent to using different weights for the timestep, while during inference, this sampling method is an adaptive step size adjustment, that is, using small-step updates with more steps in the high-noise stage and small-step updates with fewer large-step updates in the low-noise stage. In the experiment, we found that using a larger flow shift during training can bring a faster convergence speed, but has a smaller impact on the model's generation ability. Using a larger flow shift in the sampling can enhance the generation effect. As shown in the ~\figrefe{fig:flow_shift_sample}, when the flow shift is 1, it is almost impossible to generate distinguishable results. When the flow shift increases, the picture quality is significantly improved. Therefore, we used a flow shift of 1 during training and a flow shift of 17 during sampling.

\subsection{Video VAE Adaptation}
\begin{figure}[htbp]
  \centering
  \tabcolsep=3pt
  \begin{tabular}{ccccc}
    \textbf{Step 0} & \textbf{Step 100} & \textbf{Step 200} & \textbf{Step 400} & \textbf{Step 800} \\
    \includegraphics[width=0.18\textwidth]{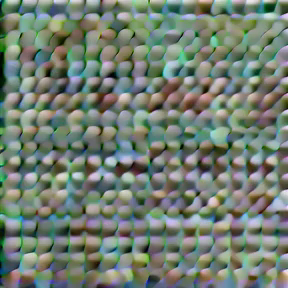} &
    \includegraphics[width=0.18\textwidth]{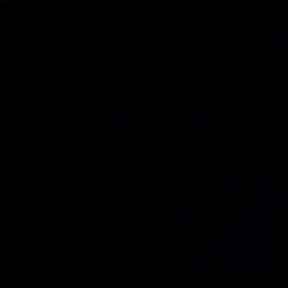} &
    \includegraphics[width=0.18\textwidth]{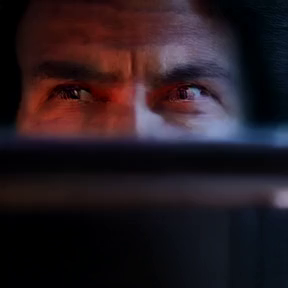} &
    \includegraphics[width=0.18\textwidth]{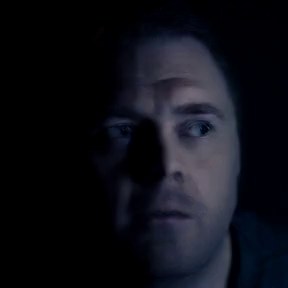} &
    \includegraphics[width=0.18\textwidth]{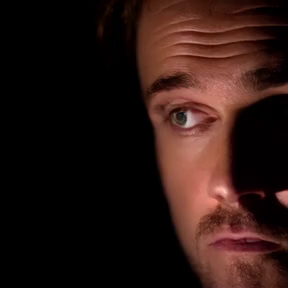} \\

    \includegraphics[width=0.18\textwidth]{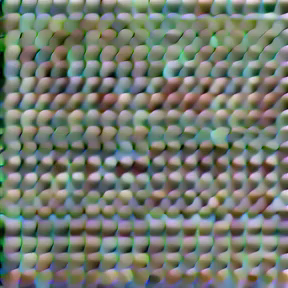} &
    \includegraphics[width=0.18\textwidth]{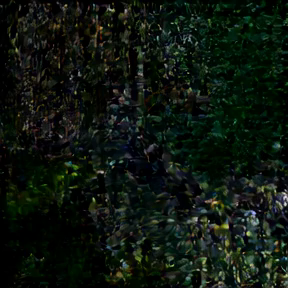} &
    \includegraphics[width=0.18\textwidth]{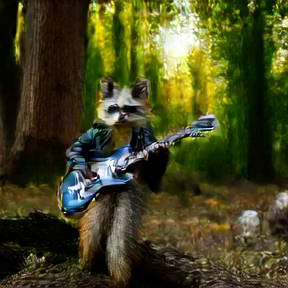} &
    \includegraphics[width=0.18\textwidth]{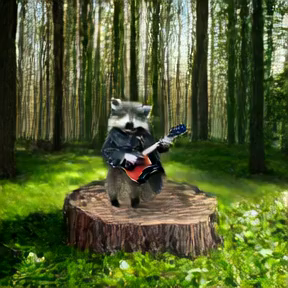} &
    \includegraphics[width=0.18\textwidth]{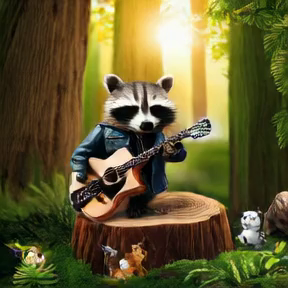} \\

    \includegraphics[width=0.18\textwidth]{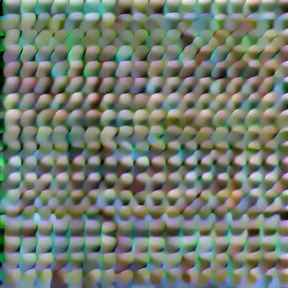} &
    \includegraphics[width=0.18\textwidth]{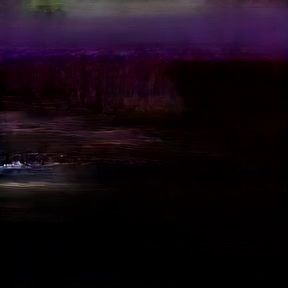} &
    \includegraphics[width=0.18\textwidth]{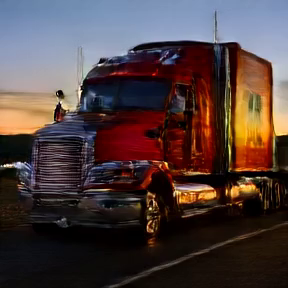} &
    \includegraphics[width=0.18\textwidth]{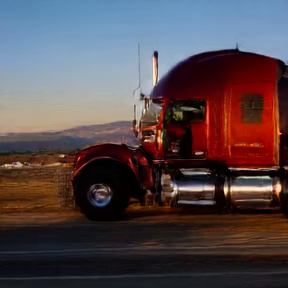} &
    \includegraphics[width=0.18\textwidth]{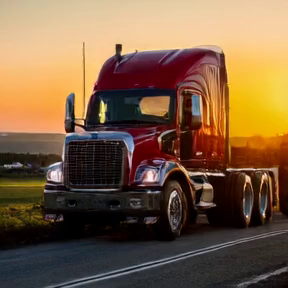} \\
  \end{tabular}
  \caption{Generated images during 3D-VAE adaptation. The rapid recovery of image generation capability demonstrates that DiT of SD3.5L can quickly adapt to the new 3D-VAE.}
  \label{fig:vae_adapt}
\end{figure}

To extend the DiT of SD3.5L from image to the video generation, a critical first step involves replacing the original 2D-VAE, which is trained exclusively on images, with a 3D-VAE capable of processing both images and videos.
The adaptation process begins by retraining the DiT model on images to reestablish its fundamental image generation capabilities. As demonstrated in \figrefe{fig:vae_adapt}, the transition to 3D VAE requires only 800 training steps for the DiT to successfully adapt to the new latent space representation while maintaining its image synthesis performance.

We quantitatively evaluated this adaptation process through Fréchet Inception Distance (FID)~\cite{heusel2017gans} measurements across 10K generated images. The FID of the SD3.5L at a resolution of 512 is 12.8. When we directly replace the VAE, all the generated images are meaningless, and the FID is 294.3, which is very large. However, after only 200 steps of training, the FID rapidly dropped to 27.8 and was able to generate distinguishable images. After training for 1600 steps, FID further decreased to 13.05, which was basically close to the generation ability before switching VAE. These quantitative results align precisely with qualitative observations from the generated samples, confirming the effectiveness and efficiency of our adaptation approach.

\subsection{DiT Pre-Training}
\paragraph{Image Video Joint Training} Our training recipe incorporates images in controlled proportions for joint video-image optimization. This strategy serves two key purposes: (1) mitigating the scarcity of high-quality video datasets by leveraging the abundant visual knowledge available in static images, and (2) maintaining the model's image comprehension capabilities despite the domain shift between video and image distributions. Contrary to conventional approaches, our analysis reveals a nuanced trade-off: while image data enhances generation quality through improved spatial understanding, it simultaneously compromises temporal dynamics. This phenomenon stems from the fundamental tension between learning spatial semantics (critical for images) and temporal coherence (essential for videos). To resolve this conflict, we implement a phased training recipe. The initial phase focuses exclusively on video data to establish robust temporal representations. Subsequently, we introduce a joint training period where both modalities are processed in balanced proportions. This staged approach enables effective cross-modal knowledge transfer while minimizing the risk of representation bias toward either domain.

\paragraph{Multiple Aspect Ratios and Durations Bucketization} After the data filtering process, the videos have different aspect ratios and durations. To effectively utilize the data, we categorize the training data into buckets based on duration and aspect ratio. Specifically, we have preset seven common video aspect buckets: $\{16:9, \ 3:2, \ 4:3, \ 1:1, \ 3:4, \ 2:3, \ 9:16\}$. Temporal duration is discretized through uniform partitioning with 1-second intervals, creating a comprehensive set of duration buckets.
To address the memory constraints imposed by variable sequence lengths across buckets, we implement a dynamic batch sizing mechanism. Each bucket is assigned an empirically determined maximum batch size that optimizes memory utilization while preventing out-of-memory conditions. All data are allocated to the nearest bucket and truncated to form a batch. During training, each rank randomly pre-fetches batch data from a bucket. This random selection ensures the model is trained on varying data sizes at each step, which helps maintain model generalization by avoiding the limitations of training on a single size. The combined bucketization approach effectively balances computational efficiency with representational diversity in the training process.

\paragraph{Progressive Video Training} 
Directly training on high-resolution, long-duration video sequences often leads to difficulties in model convergence and suboptimal results. Therefore, progressive curriculum learning has become a widely adopted strategy for training text-to-video models. In \OURMODEL, we designed a progressive training strategy, starting from the text-to-image weights and following the training method of increasing the duration first and then the resolution

\begin{itemize}
    \item Stage1: low resolution, short duration stage. The model establishes the basic mapping between text and visual content, ensuring consistency and coherence in short-term actions.
    \item Stage2: low resolution, long duration stage. The model learns more complex temporal dynamics and scene changes, ensuring temporal and spatial consistency over a longer duration.
    \item Stage3: high resolution, long duration training. The model enhances video resolution and detail quality while maintaining temporal coherence and managing complex temporal dynamics.
\end{itemize}

The detailed training configuration and hyperparameter specifications are presented in \tabrefe{tab:dit_pretrain}. We found that the model primarily acquires dynamic visual representations during stage 2 of the training process. Consequently, we allocated the majority of computational resources to this critical phase, optimizing both training efficiency and model performance. 

\begin{table}[htbp]
\caption{Training configuration across different stages.}
\label{tab:dit_pretrain}
\centering
\begin{tabular}{lccccc}
\toprule
\multicolumn{1}{c}{\multirow{2}{*}{\bf Stage}} & \multicolumn{1}{c}{\multirow{2}{*}{\bf Max Shape}} & \multicolumn{2}{c}{\bf Total Batch Size } & \multicolumn{1}{c}{\multirow{2}{*}{\bf Learning Rate}} & \multicolumn{1}{c}{\multirow{2}{*}{\bf Training Steps}} \\
\cmidrule(lr){3-4} &  &  {\bf Image} & {\bf Video} & & \\
\midrule
\textbf{VAE Adaption} & $1\times288\times288$ & 4096 & - & $1\times e^{-4}$ & 5k \\
\textbf{Stage 1} & $29\times288\times288$ & 4096 & 2048 & $1\times e^{-4}$ & 40k \\
\textbf{Stage 2} & $125\times288\times288$ & 4096 & 512 & $1\times e^{-4}$ & 60k \\
\textbf{Stage 3} & $125\times432\times768$ & 2048 & 256 & $5\times e^{-5}$ & 30k \\
\textbf{SFT} & $125\times432\times768$ & - & 256 & $1\times e^{-5}$ & 5k \\
\bottomrule
\end{tabular}
\end{table}

\section{Post-Training}
After pretraining on large-scale data, the diffusion model demonstrates the capability to generate diverse and creative videos. However, several critical challenges remain in the generated videos, including text-video misalignment, deviations from human aesthetic preferences. To enhance both the visual fidelity and motion dynamics of generated videos, we employ post-training on the pretrained model. Specifically, we conduct SFT on a high-quality subset of the pretrain data, followed by RLHF to further refine visual quality.

\subsection{Supervised Fine-Tuning}
For the data in the pre-training stage, we use a loose filtering threshold, which enables us to retain a large amount of training data, but reduces the data quality in the pre-training stage. On the other hand, due to the need for multi-duration training, the duration of the video will be truncated, resulting in inconsistency between the actual training video and the caption. These problems have led to the low quality of the videos generated by the pre-trained model and poor alignment with the prompts. Therefore, in the SFT stage, we use a high-quality subset of the pre-training data and more consistent captions to improve the performance of the model.

\paragraph{Data} As mentioned in the Section~\ref{sec:post-data}, we use stricter filtering criteria in the two dimensions of aesthetic quality and motion dynamics to filter out a high-quality subset of 1 million from the pre-training data. Furthermore, we pre-crop and truncate the video to ensure that the training video is completely aligned with the caption. At the same time, each video is marked with both short and detailed captions, and one of the captions is randomly selected during SFT.

\paragraph{Implementation Details} The SFT stage uses a smaller learning rate (10\% of pre-training learning rate) with gradient clipping to avoid catastrophic forgetting. To maintain generative diversity, we regularize the training with KL-divergence penalties against the original pre-trained model's outputs. This supervised fine-tuning stage typically requires only 5\% of the computational cost of pre-training while yielding significant improvements in video quality. 

\subsection{Reinforcement Learning from Human Feedback}
\begin{figure}[htbp]
    \centering
    \includegraphics[width=1.0\linewidth]{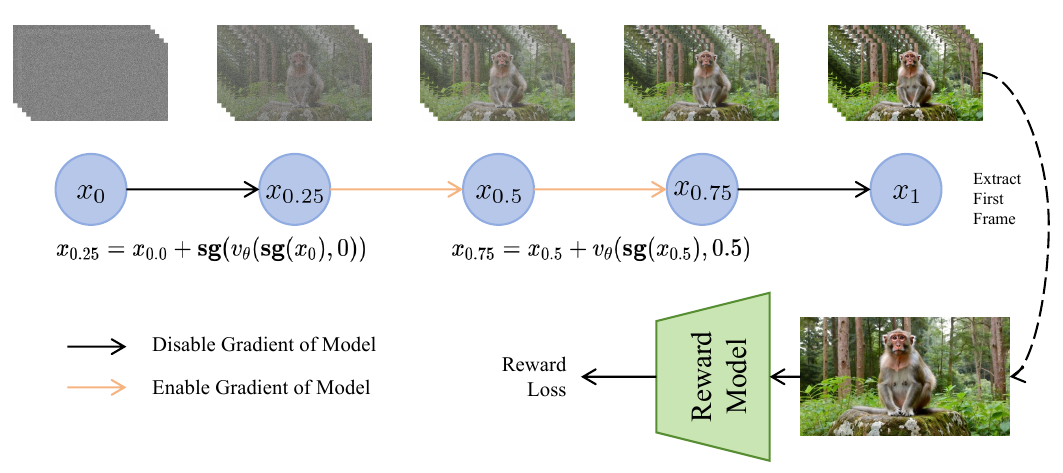}
    \caption{Illustration of diffusion RLHF. An iteration in the flow matching sampling can be written as $x_{t_{k+1}} = x_{t_k} + v_\theta(x_{t_k}, t_k)$. High-order differentiation can be avoided by stopping the gradient on the input of the model. For the random $k$ steps in the sampling process, the model gradient is enabled. At this time, $x_1$ contains the gradient and the parameters of these $k$ steps can be optimized through backpropagation.
    }
    \label{fig:diffusion_rlhf}
\end{figure}

RLHF for diffusion model aims to optimize a conditional distribution $p_\theta(x_1\mid c)$ such that the reward model $r(c, x_1)$ defined on it is maximized while regularizing the KL-divergence from a reference model $p_\textrm{ref}$. More specifically, RLHF optimizes a model $p_\theta$ to maximize the following objective:
\begin{equation}
\max_{p_\theta} \mathbb{E}_{c\sim\mathcal{D}_c, x_1\sim p_\theta(x_1 \mid c)}[r(c, x_1)] - \beta \mathbb{D}_{\mathrm{KL}}[p_\theta(x_1\mid c)\parallel p_\mathrm{ref}(x_1 \mid c)]
\label{eq:loss-rlhf}
\end{equation}
where the hyperparameter $\beta$ controls KL-regularization strength. 
RLHF training requires the sampled $x_1$ be differentiable. As shown in the \eqrefe{eq:flow-matching-sampling}, the sampling result $x_1$ of flow matching can be written as the sum of the outputs of the model at each step.
\begin{equation}
x_1 = x_0 + \frac{1}{N}\sum_{i=0}^{N-1}v_\theta(x_{\frac{i}{N}}, \frac{i}{N})
\label{eq:flow-matching-sampling}
\end{equation}
Inspired by \citet{DRTune}, we enable the gradient of the model to make $x_1$ differentiable and maximize the reward through backpropagation.
Instead of enabling the gradient at equal intervals, we randomly sample $k$ steps carrying gradient from the timesteps of inference as shown in the \figrefe{fig:diffusion_rlhf}. Considering that the model for video generation is large and the sequence is long while the memory of the NPUs is limited, we ignore the KL constraint for the sake of computational efficiency. 

\paragraph{Data}
For the RLHF training, we constructed a high-quality prompts set through a rigorous multi-stage process. First, we randomly selected 80,000 prompts from our SFT dataset, ensuring balanced coverage across different content categories and prompt styles. The SFT dataset itself was carefully curated from diverse sources including: (1) professionally created content descriptions, (2) user-generated prompts from creative platforms, and (3) synthetic prompts generated by language models with human verification.

\paragraph{Reward Models} Initially, we employed VideoAlign~\cite{VideoAlign}, a reward model based on the visual language model (VLM), to evaluate generated videos across three key dimensions: visual quality (VQ), motion quality (MQ), and text alignment (TA). During training, we observed that while the reward scores improved steadily, the qualitative performance of the generated videos showed no significant enhancement. Further analysis revealed that the VLM-based reward model introduced an unintended bias: instead of holistically optimizing video quality, the model disproportionately prioritized literal prompt adherence, leading to suboptimal generations. We choose MPS~\cite{mps}, a CLIP-based model, as our visual quality reward model. It can effectively enhance the visual quality of the video.

\paragraph{Implementation Details} The process of generating initial videos $x_1$ from noise through diffusion models proves computationally intensive, particularly for extended video sequences. A complete iterative generation cycle typically requires tens of minutes to execute, while the subsequent decoding from latent space to pixel space demands substantial video memory resources. To address these efficiency challenges during the RLHF stage, we implemented two key optimizations: (1) reducing the generated video length from 125 frames to 29 frames, which dramatically decreases generation time; and (2) employing a selective decoding strategy that leverages the causal properties of VAE architecture. Specifically, we decode only the first frame and utilize MPS to enhance the perceptual quality of generated images while maintaining computational efficiency. These modifications collectively improve the training pipeline's throughput without compromising output quality.


\section{Infrastructure}
This section presents the infrastructure designed to enable efficient and scalable training of the text-to-video model. Our training is based on a cluster consisting of 256 NPUs.
To address the substantial memory and computational demands of large-scale video generation, we implement a series of optimization strategies. These include: (1) architectural separation between the feature encoder and training modules to reduce memory overhead; (2) advanced memory optimization techniques to maximize hardware utilization; and (3) a 3D parallelism strategy that combines data, param, and sequence parallelism. Through the synergistic application of these methods, we achieve stable training of an 8-billion parameters video generation model within the 64~GB memory constraints.

\subsection{Hardware}
Our training setup employs a cluster consisting of 32 nodes, each node equipped with 8 NPUs and the torch\_npu 2.1.0 deep learning framework. To optimize distributed training performance, we leverage the Collective Communication Library (HCCL), a high-performance communication library that enables efficient multi-node synchronization with high bandwidth and low latency. The interconnect architecture delivers exceptional bandwidth capabilities: each node supports a one-to-one communication bandwidth of 56~GB/s and an aggregate one-to-many bandwidth of 392~GB/s. The NPU feature 64~GB of High Bandwidth Memory (HBM) per card and deliver a computational throughput of 320~TFLOPS when operating with BF16/FP16 mixed-precision. This hardware configuration provides the necessary computational density and communication bandwidth for large-scale distributed training workloads.

\subsection{Asynchronous Encode Server}
\begin{figure}[htbp]
    \centering
    \includegraphics[width=1.0\linewidth]{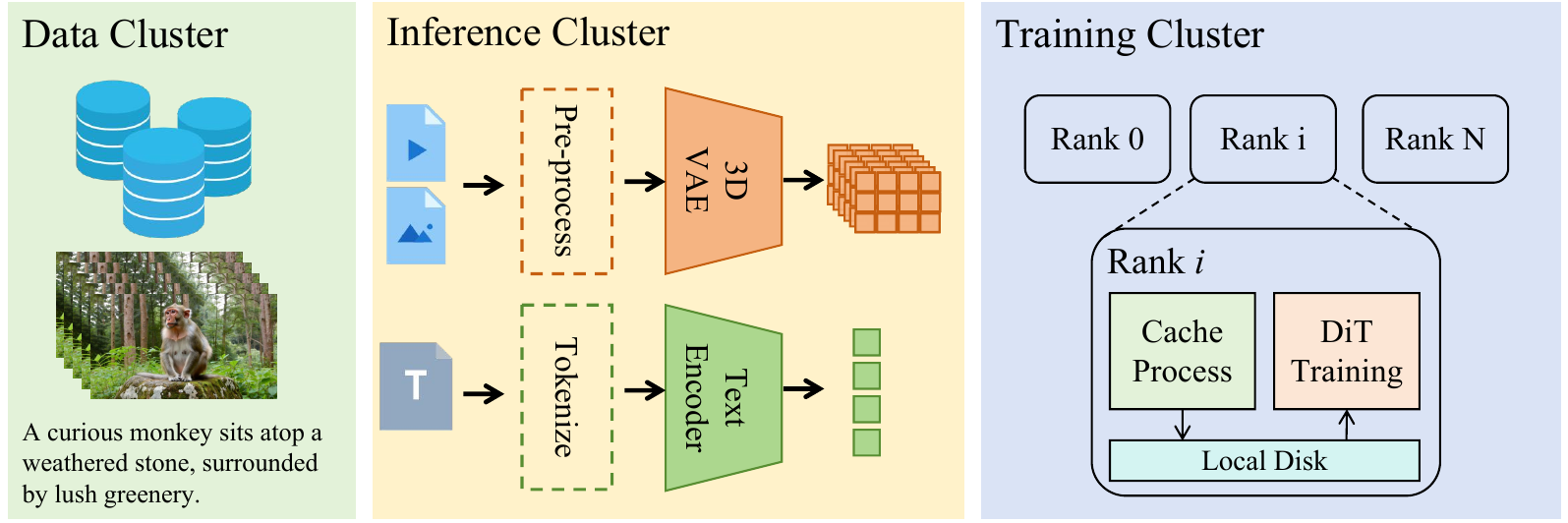}
    \caption{Asynchronous Encoding Server Architecture. The feature extraction and model training processes are distributed across separate computing clusters. Data and feature transfers are handled asynchronously via network requests to optimize efficiency and resource utilization.}
    \label{fig:syncserver}
\end{figure}

Since the text-to-video diffusion model performs denoising in the latent space, it requires both the VAE and text encoder to extract latent features and text embeddings, respectively. Although this inference stage involves no gradient computation, the T5-XXL~\cite{t5xxl} text encoder and VAE encoder still demand significant memory resources. In SD3.5L~\cite{sd3}, feature extraction is performed offline, which inevitably consumes substantial storage space. To address these resource constraints, we propose a decoupled architecture that separates computing resources for inference and training, as shown in the \figrefe{fig:syncserver}. Specifically, the VAE and text encoder operate on dedicated inference clusters, continuously generating processed input data (including image and video latent features and corresponding text embeddings) for DiT training. The training cluster then retrieves these pre-computed features via network transmission and focuses exclusively on the training process.

\paragraph{Batch Buffer} For each rank, we employ a dedicated process to manage a data buffer. During the initial phase, while the buffer has not reached its full capacity, the process iteratively fetches new batch data from the server and stores it locally. This design enables training nodes to directly access the required features from their local storage, eliminating the need for frequent remote data retrieval during training. To ensure synchronization across all ranks, we utilize the training step count as a unified identifier, which guarantees consistent coordination between data loading and model training processes.

\subsection{Memory Optimization}
\paragraph{Attention optimization} As the sequence length grows, the computational overhead of attention becomes the primary bottleneck in training. We leverage \texttt{npu\_fusion\_attention} to accelerate attention calculations, achieving significant improvements in computational efficiency. Additionally, this optimization substantially reduces memory consumption, enabling the training of longer sequences within constrained hardware resources.

\paragraph{Gradient Checkpointing} Gradient checkpointing is a memory optimization technique that trades computational overhead for reduced memory requirements. The method consists of three key steps: (1) selectively designating specific layers or blocks for recomputation, (2) releasing intermediate activations during the forward pass, and (3) recalculating these activations as needed during the backward pass. This approach significantly decreases memory consumption during training, enabling the processing of larger models or batch sizes within limited memory constraints.

\subsection{Parallelism Strategies}
The substantial model size and the exceptionally long sequence length (exceeding 116K tokens for the longest sequence) necessitates the implementation of advanced parallelism strategies to ensure computationally efficient training. To address this challenge, we adopt 3D parallelism, a scalable approach that combines data parallelism, tensor (parameter) parallelism, and sequence parallelism. This strategy enables efficient distribution of computational workloads across all three critical dimensions, including data batches, model parameters, and long sequences, thereby optimizing memory usage and computational throughput.

\paragraph{Sequence Parallelism(SP)} 
Sequence Parallelism is a technique that partitions input tensors along the sequence dimension, enabling independent processing of layers such as LayerNorm to eliminate redundant computations and reduce memory overhead~\citep{korthikanti2023reducing,li2021sequence,jacobs2023deepspeed}. This approach also facilitates efficient handling of variable-length sequences by supporting dynamic padding for non-conforming inputs. Our implementation leverages Ulysses~\citep{jacobs2023deepspeed}, a framework that shards input samples across the sequence parallel group at the beginning of the training loop. During attention computation, Ulysses employs all-to-all communication to distribute query, key, and value tensors such that each worker processes the full sequence length but only a subset of attention heads. Following parallel computation, a second all-to-all operation reassembles the outputs, reconstructing both the attention heads and the original sequence shards. To ensure robustness, our implementation includes engineering optimizations for input sequences that do not inherently meet SP requirements, such as automatic padding to maintain computational efficiency.

\paragraph{Fully Sharded Data Parallelism}  
\citet{zhao2023pytorch} introduces Fully Sharded Data Parallelism (FSDP), a memory-efficient distributed training approach that partitions model parameters, gradients, and optimizer states across data-parallel ranks. Unlike Distributed Data Parallelism (DDP), which relies on all-reduce for synchronization, FSDP employs all-gather for parameter retrieval and reduce-scatter for gradient aggregation. These operations are strategically overlapped with forward and backward computations, reducing communication overhead and improving training efficiency. In our implementation, we leverage the \texttt{HYBRID\_SHARD} strategy, which combines intra-group parameter sharding (\texttt{FULL\_SHARD}) with inter-group parameter replication. This hybrid approach effectively implements hierarchical data parallelism: while \texttt{FULL\_SHARD} optimizes memory usage within each shard group, cross-group replication enhances computational throughput. By localizing all-gather and reduce-scatter operations within shard groups, this method significantly reduces inter-rank communication costs compared to global synchronization schemes.

\begin{figure}[htbp]
    \begin{subfigure}{\textwidth}
        \centering
        \caption*{CogVideoX-5B}
        \includegraphics[width=\textwidth]{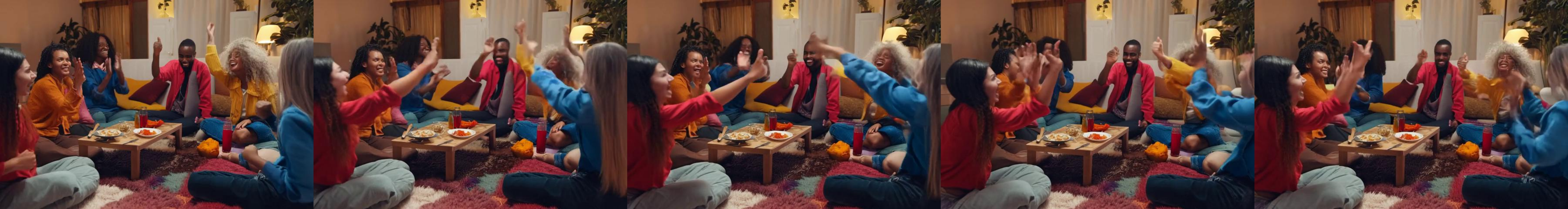}
    \end{subfigure}
    \begin{subfigure}{\textwidth}
        \centering
        \caption*{HunyuanVideo-13B}
        \includegraphics[width=\textwidth]{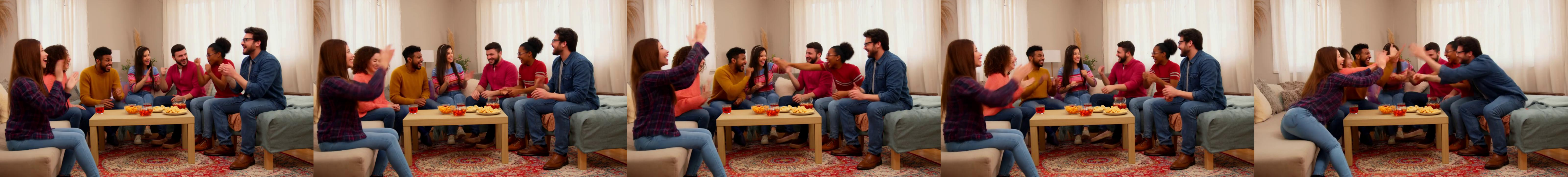}
    \end{subfigure}
    \begin{subfigure}{\textwidth}
        \centering
        \caption*{Wan2.1-14B}
        \includegraphics[width=\textwidth]{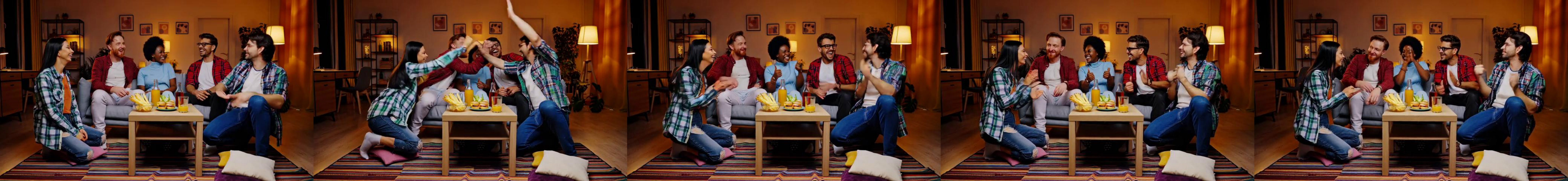}
    \end{subfigure}
    \begin{subfigure}{\textwidth}
        \centering
        \caption*{MINIMAX Hailuo}
        \includegraphics[width=\textwidth]{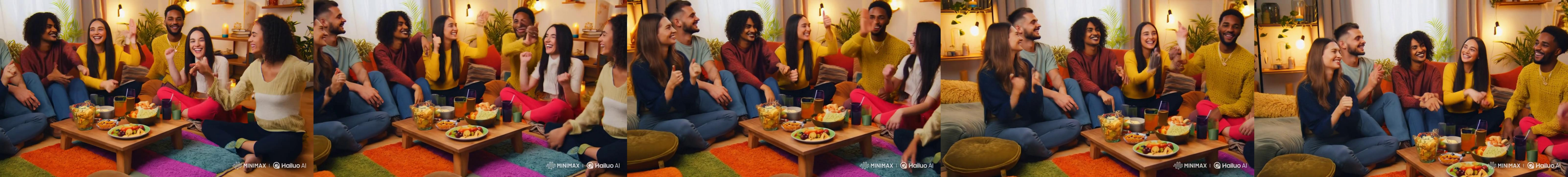}
    \end{subfigure}
    \begin{subfigure}{\textwidth}
        \centering
        \caption*{Vidu}
        \includegraphics[width=\textwidth]{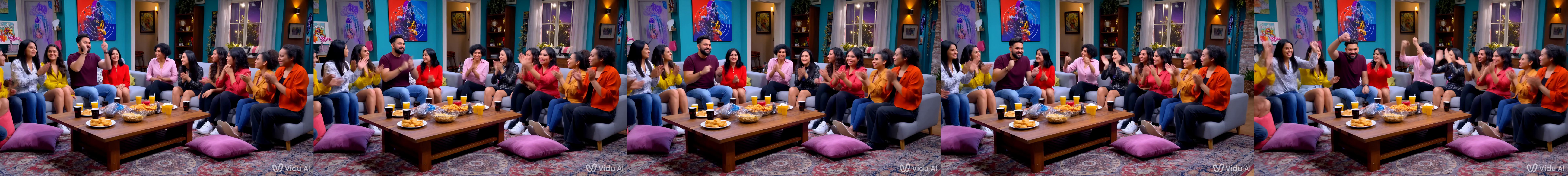}
    \end{subfigure}
    \begin{subfigure}{\textwidth}
        \centering
        \caption*{Kling 1.6}
        \includegraphics[width=\textwidth]{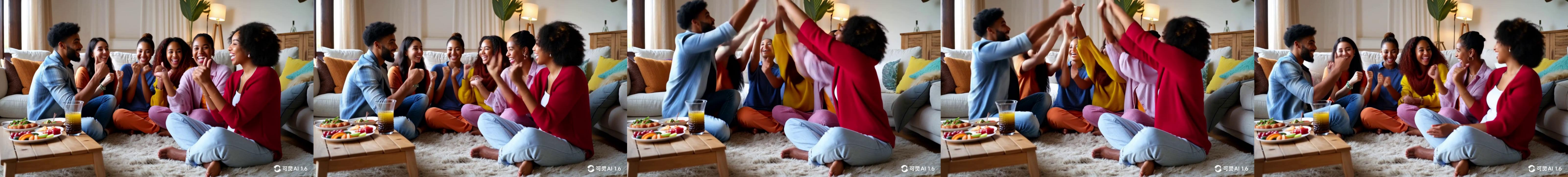}
    \end{subfigure}
    \begin{subfigure}{\textwidth}
        \centering
        \caption*{Jimeng PixelDance 2.0 Pro}
        \includegraphics[width=\textwidth]{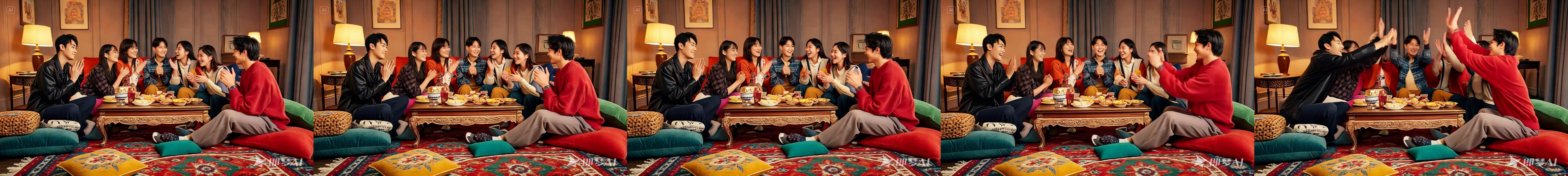}
    \end{subfigure}
    \begin{subfigure}{\textwidth}
        \centering
        \caption*{\OURMODEL~-8B}
        \includegraphics[width=\textwidth]{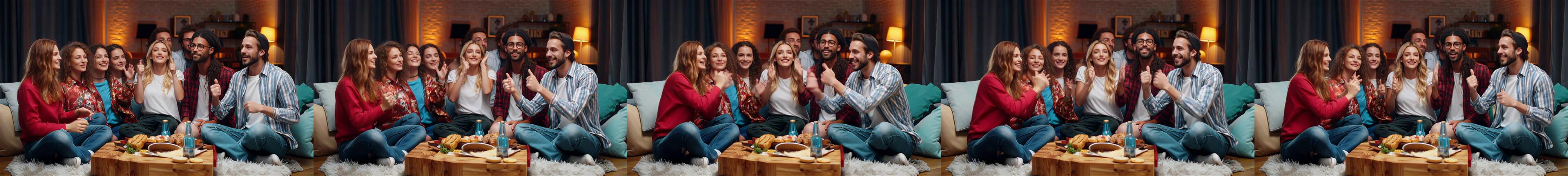}
    \end{subfigure}
    \caption{Qualitative comparison between \OURMODEL~and other T2V models. Prompt: \textit{A lively group of friends, diverse in appearance and style, gather in a cozy, warmly lit living room, filled with laughter and camaraderie. They sit on a plush, colorful rug, surrounded by soft cushions and a low wooden table adorned with snacks and drinks. Each friend, dressed in casual, vibrant attire, expresses agreement through enthusiastic hand gestures, such as thumbs up, high-fives, and fist bumps. Their faces light up with genuine smiles and nods, reflecting a shared understanding and mutual support. The room's ambiance, with its soft lighting and eclectic decor, enhances the sense of warmth and friendship.}}
    \label{fig:compare_model}
\end{figure}

\section{Performance}
\subsection{Qualitative Visualizations}
\begin{figure}[htbp]
    \begin{subfigure}{\textwidth}
        \centering
        \includegraphics[width=\textwidth]{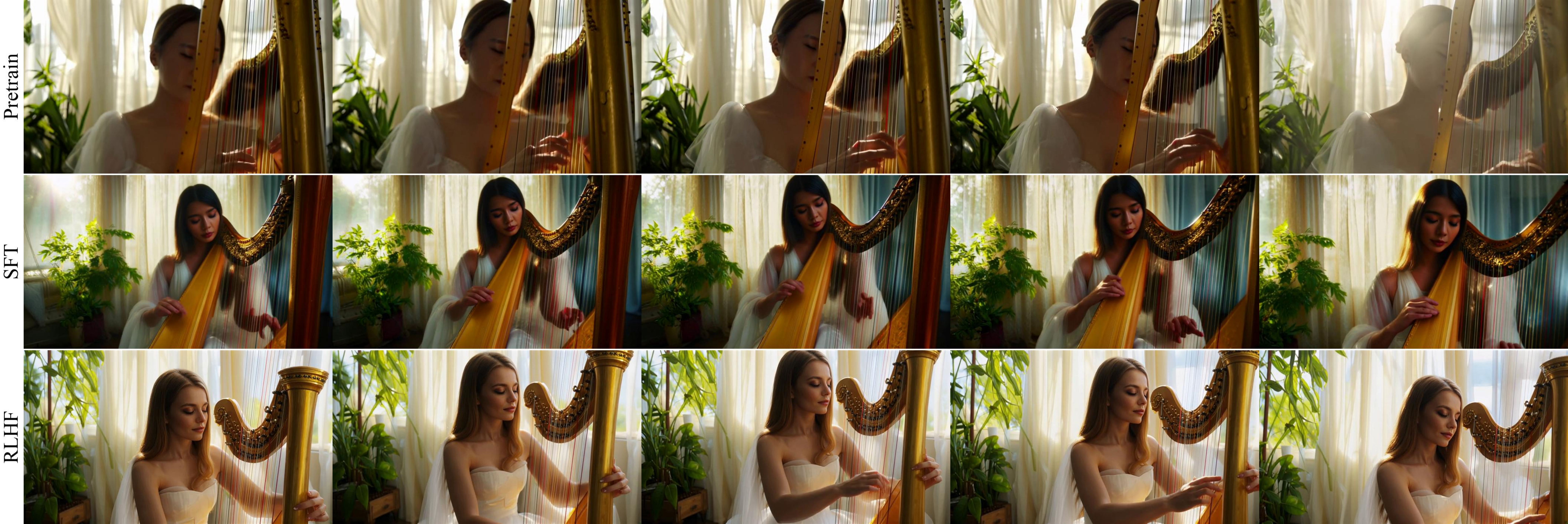}
        \caption{\textit{A serene individual, dressed in a flowing white gown, sits gracefully in a sunlit room adorned with lush green plants and soft, billowing curtains. Their fingers delicately pluck the strings of a golden harp, producing ethereal melodies that fill the air. The camera captures close-ups of hands, showcasing the intricate movements and the harp's ornate details. Sunlight filters through the window, casting a warm glow on serene face, eyes closed in deep concentration. The scene transitions to a wider shot, revealing the tranquil ambiance of the room, with the gentle sway of the curtains and the soft rustle of leaves enhancing the peaceful atmosphere.}}
    \end{subfigure}
    \begin{subfigure}{\textwidth}
        \centering
        \includegraphics[width=\textwidth]{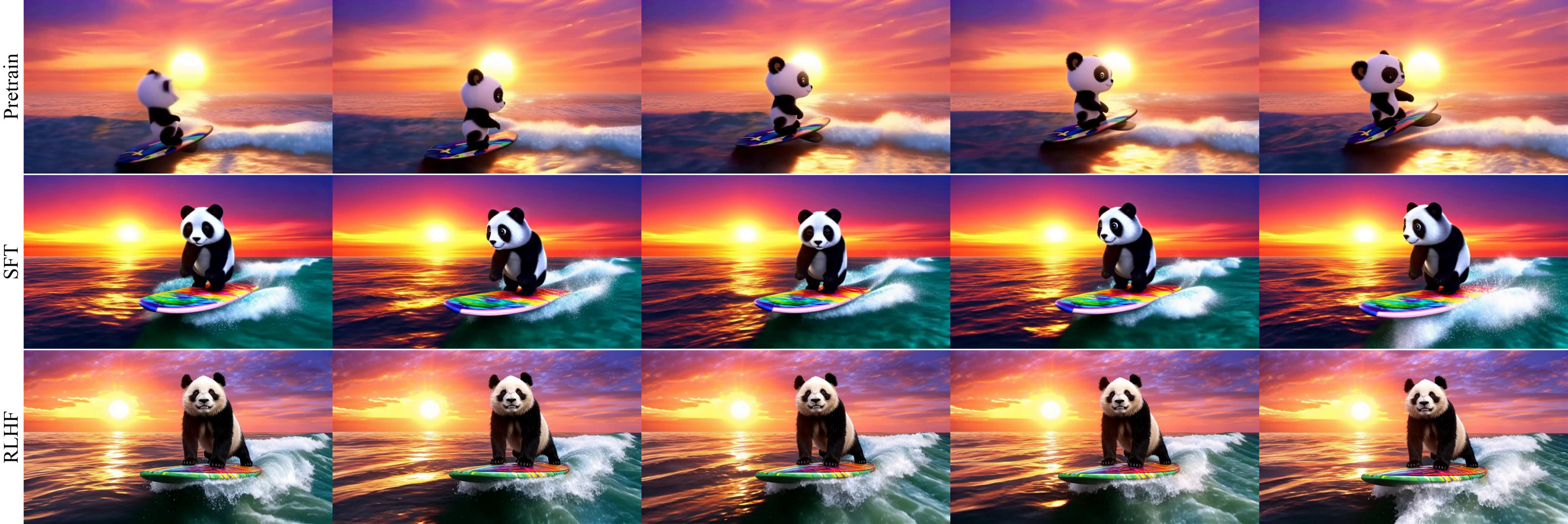}
        \caption{\textit{A playful panda stands confidently on a surfboard, riding gentle waves in the ocean during a breathtaking sunset. The sky is ablaze with hues of orange, pink, and purple, casting a warm glow on the water. The panda, with its black and white fur glistening in the golden light, balances effortlessly, its eyes wide with excitement. The surfboard, painted in vibrant colors, cuts through the shimmering waves, leaving a trail of sparkling droplets. In the background, the sun dips below the horizon, creating a serene and magical atmosphere, as the panda enjoys its unique adventure amidst the tranquil sea.}}
    \end{subfigure}
    \begin{subfigure}{\textwidth}
        \centering
        \includegraphics[width=\textwidth]{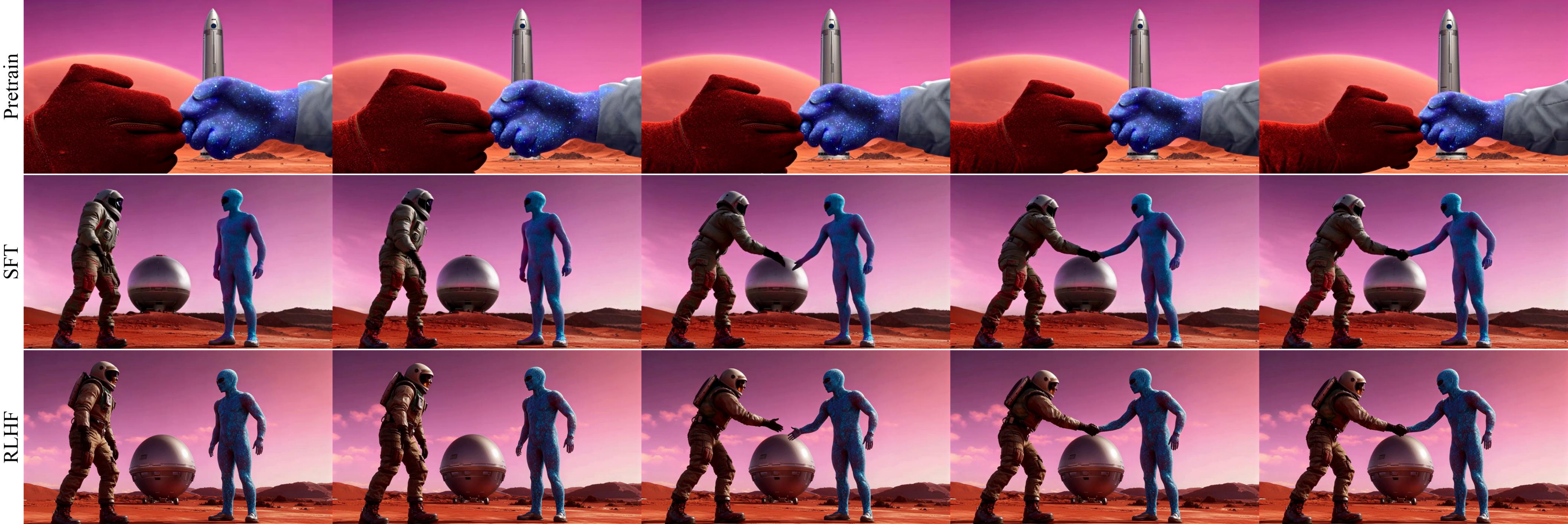}
        \caption{\textit{A suited astronaut, with the red dust of Mars clinging to boots, reaches out to shake hands with an alien, their skin a shimmering blue, under the pink-tinged sky of the fourth planet. In the background, a sleek silver rocket, stands tall, its engines powered down, as the two representatives of different worlds exchange a historic greeting amidst the desolate beauty of the Martian landscape.}}
    \end{subfigure}
    \caption{Qualitative comparison of \OURMODEL~at different stages.}
    \label{fig:compare_stage}
\end{figure}

For intuitive comparisons, we conduct qualitative assessments and present sampled results in \figrefe{fig:compare_model}. The evaluation includes open-source models CogVideoX-5B~\cite{yang2024cogvideox}, HunyuanVideo-13B~\cite{kong2024hunyuanvideo}, Wan-14B~\cite{wang2025wan}, as well as closed-source products, including MINIMAX Hailuo~\cite{hailuo}, Vidu~\cite{vidu}, Kling1.6~\cite{kling}, and Jimeng PixelDance 2.0~\cite{jimeng}. The ~\figrefe{fig:compare_model} shows that, compared with other open-source models, \OURMODEL~can achieve better visual quality and be comparable to some closed-source models. This is because RLHF aligns with human preferences, significantly enhancing the visual quality of the video. We present the visualization results of the Pre-train, SFT, and RLHF stages respectively in \figrefe{fig:compare_stage}. The SFT stage enhances the semantic alignment between the video and the prompt, while RLHF improves the visual quality and details, particularly in terms of human preference. More video samples are shown on the project page.

\subsection{Quantitative Analysis}
For quantitative analysis, we evaluate \OURMODEL~against the SOTA text-to-video models on VBench~\citep{vbench}, a benchmark for comprehensively assessing the quality of video generation from 16 dimensions. Vbench provides two types of prompts: One is short and simple prompts, with an average of 7.6 words. Another type is long and detailed prompts enhanced by chatGPT, with an average of 100.5 words. We evaluate both types of prompts simultaneously to comprehensively measure the capabilities of \OURMODEL.

As illustrated in \tabrefe{tab:benchmark_model}, \OURMODEL~demonstrated outstanding performance in multiple evaluation dimensions and outperformed numerous open-source and commercial models in the overall score. Among all open-source models, \OURMODEL~is only inferior to Wan2.1~\cite{wang2025wan}. However, compared with Wan2.1 with 14B parameters, \OURMODEL~achieved comparable performance by using only 8B (\textasciitilde 57\% of 14B) parameters.
Notably, \OURMODEL~achieved a higher VBench overall score in long prompts, mainly because detailed prompts brought a significant improvement of 3.51 in the semantic score. This indicates that \OURMODEL~has the ability to understand complex prompts and enhance the alignment of text-video.

\begin{table}[htbp]\small
\caption{Comparison with leading T2V models on VBench\citep{vbench} in descending order of overall score. \dag~denotes models with \textbf{non-publicly} available weights. The results of other models come from official VBench leaderboard.}
\label{tab:benchmark_model}
\tabcolsep=5pt
\centering
\begin{tabular}{lcccccccc}
\toprule
\bf Models & \bf Overall & \bf \makecell{Quality\\Score} & \bf \makecell{Semantic\\Score} & \bf \makecell{Human\\Action}  & \bf Scene &  \bf \makecell{Dynamic\\Degree}  & \bf \makecell{Multiple\\Objects}  & \bf \makecell{Appear.\\Style}  \\ 
\midrule
\href{https://www.vidu.cn/}{Vidu-Q1}\dag & 87.41 & 87.28 & 87.94 & 99.60 & 67.41 & 91.85 & 93.89 & 22.37 \\
\href{https://github.com/Wan-Video/Wan2.1}{Wan2.1-14B} & 86.22 & 86.67 & 84.44 & 99.20 & 61.24 & 94.26 & 86.59 & 21.59 \\
\cellcolor{cyan!15}\textbf{\OURMODEL}(Long) & \cellcolor{cyan!15}85.14 & \cellcolor{cyan!15}86.64 & \cellcolor{cyan!15}79.12 & \cellcolor{cyan!15}96.80 & \cellcolor{cyan!15}57.38 & \cellcolor{cyan!15}83.05 & \cellcolor{cyan!15}71.41 & \cellcolor{cyan!15}23.02 \\
\href{https://github.com/Saiyan-World/goku}{Goku}\dag & 84.85 & 85.60 & 81.87 & 97.60 & 57.08 & 76.11 & 79.48 & 23.08 \\
\href{https://github.com/hpcaitech/Open-Sora}{Open-Sora 2.0} & 84.34 & 85.40 & 80.12 & 95.40 & 52.71 & 71.39 & 77.72 & 22.98 \\
\href{https://openai.com/sora/}{Sora}\dag & 84.28 & 85.51 & 79.35 & 98.20 & 56.95 & 79.91 & 70.85 & 24.76 \\
\href{https://causvid.github.io/}{CausVid}\dag & 84.27 & 85.65 & 78.75 & 99.80 & 56.58 & 92.69 & 72.15 & 24.27 \\
\cellcolor{cyan!15}\textbf{\OURMODEL}(Short) & \cellcolor{cyan!15}84.11 & \cellcolor{cyan!15}86.23 & \cellcolor{cyan!15}75.61 & \cellcolor{cyan!15}89.60 & \cellcolor{cyan!15}44.02 & \cellcolor{cyan!15}79.26 & \cellcolor{cyan!15}74.58 & \cellcolor{cyan!15}21.21 \\
\href{https://lumalabs.ai/dream-machine}{Luma}\dag & 83.61 & 83.47 & 84.17 & 96.40 & 58.98 & 44.26 & 82.63 & 24.66 \\
\href{https://github.com/aigc-apps/EasyAnimate}{EasyAnimate 5.1} & 83.42 & 85.03 & 77.01 & 95.60 & 54.31 & 57.15 & 66.85 & 23.06 \\
\href{https://hailuoai.video}{Hailuo-Video-01}\dag & 83.41 & 84.85 & 77.65 & 92.40 & 50.68 & 64.91 & 76.04 & 20.06 \\
\href{https://klingai.kuaishou.com/}{Kling 1.6}\dag & 83.40 & 85.00 & 76.99 & 96.20 & 55.57 & 62.22 & 63.99 & 20.75 \\
\href{https://github.com/Tencent/HunyuanVideo}{HunyuanVideo} & 83.24 & 85.09 & 75.82 & 94.40 & 53.88 & 70.83 & 68.55 & 19.80 \\
\href{https://runwayml.com/research/introducing-gen-3-alpha}{Gen-3}\dag & 82.32 & 84.11 & 75.17 & 96.40 & 54.57 & 60.14 & 53.64 & 24.31 \\
\href{https://github.com/THUDM/CogVideo}{CogVideoX-5B} & 81.61 & 82.75 & 77.04 & 99.40 & 53.20 & 70.97 & 62.11 & 24.91 \\
\href{https://pika.art}{Pika-1.0}\dag & 80.69 & 82.92 & 71.77 & 86.20 & 49.83 & 47.50 & 43.08 & 22.26 \\
\href{https://github.com/AILab-CVC/VideoCrafter}{VideoCrafter-2.0} & 80.44 & 82.20 & 73.42 & 95.00 & 55.29 & 42.50 & 40.66 & 25.13 \\
\href{https://github.com/guoyww/AnimateDiff}{AnimateDiff-V2} & 80.27 & 82.90 & 69.75 & 92.60 & 50.19 & 40.83 & 36.88 & 22.42 \\
\href{https://huggingface.co/hpcai-tech/OpenSora-STDiT-v3}{OpenSora 1.2} & 79.23 & 80.71 & 73.30 & 85.80 & 42.47 & 47.22 & 58.41 & 23.89 \\
\href{https://github.com/showlab/Show-1}{Show-1} & 78.93 & 80.42 & 72.98 & 95.60 & 47.03 & 44.44 & 45.47 & 23.06 \\
\href{https://github.com/mira-space/Mira}{Mira} & 71.87 & 78.78 & 44.21 & 63.80 & 16.34 & 60.33 & 12.52 & 21.89 \\
\bottomrule
\end{tabular}
\end{table}

Furthermore, to ensure the robust evaluation, we use the end-to-end visual language model based evaluator VideoAlign\cite{VideoAlign} to complement the VBench evaluation criteria. VideoAlign provides fine-grained analysis across three dimensions of video generation: visual quality (VQ) quantifies frame-level clarity and structural fidelity; motion quality (MQ) evaluates temporal coherence and physical plausibility of movements; and text alignment (TA) measures semantic consistency between the generated video and its textual prompt. We use both Vbench\cite{vbench} and VideoAlign\cite{VideoAlign} to measure the impact of the Pre-train, SFT, RLHF stages on the generated videos. 

\begin{table}[htbp]
\caption{Improvements of \OURMODEL~across stages on the experimental model. The {\color{blue}blue} represents win rates against the pretrain model.}
\label{tab:benchmark_stage}
\centering
\begin{tabular}{cccccccc}
\toprule
    \multicolumn{1}{c}{\multirow{2}{*}{\bf Stage}} & \multicolumn{3}{c}{\textbf{Vbench}\cite{vbench} } & \multicolumn{3}{c}{\textbf{VideoAlign}\cite{VideoAlign} } \\
    \cmidrule(lr){2-4} \cmidrule(lr){5-8} & {\bf Quality} & {\bf Semantic} &  {\bf Overall} & {\bf VQ} & {\bf MQ} &  {\bf TA} & {\bf Overall} \\
\midrule
\textbf{Pretrain} & 85.11 & 78.29 & 83.74 & -0.2160 & 0.2494 & 0.8737 & 0.9071 \\
\textbf{SFT} & 85.24 & 79.73 & 84.13 & \makecell{-0.2080\\ [-0.8ex]\scriptsize \color{blue}50.26\%} & \makecell{0.2443\\ [-0.8ex]\scriptsize\color{blue} 46.40\%} & \makecell{1.0145\\ [-0.8ex]\scriptsize\color{blue} 61.31\%} & \makecell{1.0508\\ [-0.8ex]\scriptsize\color{blue} 56.77\%} \\
\textbf{RLHF} & 85.85 & 80.02 & 84.68 & \makecell{0.3034\\ [-0.8ex]\scriptsize\color{blue} 89.38\%} & \makecell{0.4339\\ [-0.8ex]\scriptsize\color{blue} 65.96\%} & \makecell{1.1599\\ [-0.8ex]\scriptsize\color{blue} 76.48\%} & \makecell{1.8972\\ [-0.8ex]\scriptsize\color{blue} 85.62\%} \\
\bottomrule
\end{tabular}
\end{table}

The results in \tabrefe{tab:benchmark_stage} demonstrate progressive performance improvements across successive training stages, with each phase contributing distinct enhancements to different evaluation metrics.
During the SFT stage, the model shows a significant improvement in TA performance, achieving a 61.31\% win rate from 0.8737 to 1.0145. This improvement is attributed to the utilization of filtered video samples from the pre-training dataset accompanied by more precise and comprehensive captions. Notably, while TA shows substantial gains, VQ and MQ metrics remain stable, maintaining performance levels comparable to the baseline. This observation suggests that the SFT process primarily optimizes text-video alignment through improved caption quality rather than affecting the visual quality.
The RLHF stage produces more comprehensive improvements across all evaluation dimensions. Most remarkably, this stage yields an 89.38\% win rate enhancement in VQ (from -0.2160 to 0.3034) and an 85.62\% win rate improvement in overall score (from 0.9071 to 1.8972). These substantial gains can be attributed to RLHF's optimization of the complete diffusion process, which systematically aligns the denoised $x_1$ output with human preference patterns through iterative feedback mechanisms.

\subsection{User Study}
Despite the use of multi-dimensional evaluation schemes in Vbench and VideoAlign, there is still a gap between model-based evaluation and human evaluation. 
Therefore, we conducted a user study between \OURMODEL~and three other mainstream open-source models, namely CogvideoX-5B~\cite{yang2024cogvideox}, HunyuanVideo-13B~\cite{kong2024hunyuanvideo}, and WanX2.1-14B~\cite{wang2025wan} . 
Specifically, our evaluation examines four critical dimensions of video generation quality:
\begin{itemize}
    \item \textbf{Perceptual Quality}: Assesses fundamental video characteristics including temporal consistency, frame-to-frame stability, motion fluidity, and overall aesthetic coherence.
    \item \textbf{Instruction Following}: Accuracy in matching textual prompts for elements, quantities, and features
    \item \textbf{Physical Simulation}: Realism in motion, lighting, and physical interactions
    \item \textbf{Visual Quality}: Quantifies production-grade quality metrics including resolution fidelity, color accuracy, noise levels, and artifact presence.
\end{itemize}

\paragraph{Evaluation Metrics} We use the Good-Same-Bad (GSB) ratio to quantitatively represent the differences in human preferences between the evaluation model, namely \OURMODEL, and the comparison model. Given a prompt and videos generated respectively by the evaluation model and the comparison model, human annotators need to make a choice based on their personal preferences: the evaluation model is better ("\underline{G}ood"), the two are about the same ("\underline{S}ame"), and the comparison model is better ("\underline{B}ad"). The GSB ratio is calculated as:
\begin{equation}
\text{GSB Ratio} = \frac{G + S}{B + S}
\end{equation}
A ratio greater than 1 indicates superior performance of the evaluated model, while a ratio below 1 suggests the comparison baseline performs better. For instance, when assessing 10 sample videos with results of 7 better, 2 equivalent, and 1 worse cases, the resulting 7:2:1 ratio yields a GSB value of 3.0, clearly indicating performance superiority.

\begin{figure}[htbp]
    \centering
    \includegraphics[width=\textwidth]{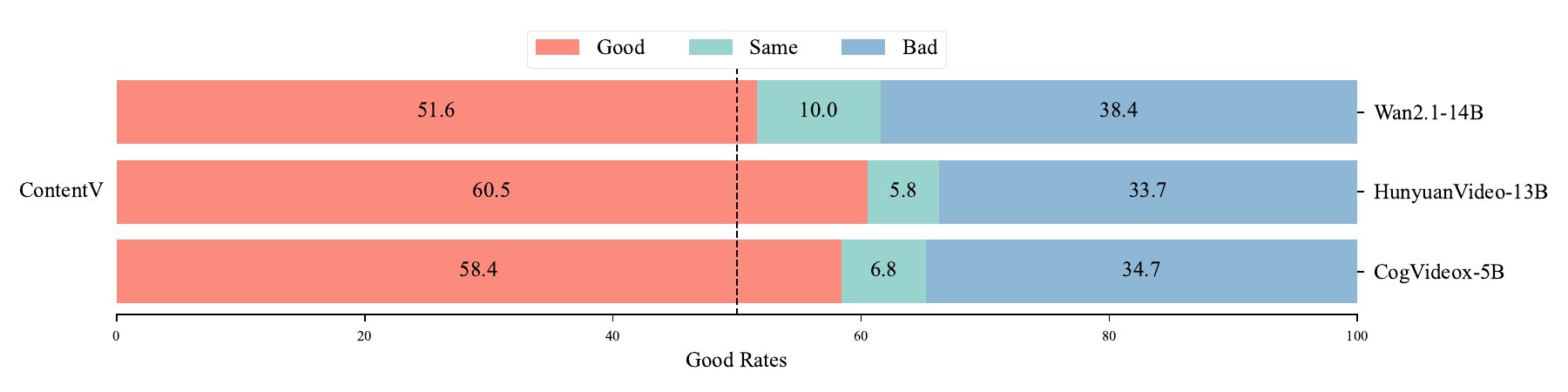}
    \caption{User studies of \OURMODEL~against open-source SOTA T2V models. Compared to CogVideoX, HunyuanVideo, and Wan2.1, \OURMODEL~achieves a GSB ratio of 1.57, 1.68, and 1.30, respectively.}
    \label{fig:human_eval}
\end{figure}

We invite five human annotators to comprehensively consider and select the preferred videos. The comparative GSB results are presented in \figrefe{fig:human_eval}. Consistent with the model-based evaluation, \OURMODEL~demonstrates superior performance over both CogVideoX and HunyuanVideo across overall evaluation dimensions. An interesting observation emerges from the comparison between metric-based and human evaluations: while \OURMODEL~showed lower scores than Wan2.1 on VBench, it achieved better human preference ratings. This discrepancy suggests that the VBench evaluation criteria may emphasize different aspects of video quality compared to human perception. We hypothesize that human evaluators place greater emphasis on holistic factors such as natural motion, temporal coherence, and aesthetic quality, which may not be fully captured by automated metric systems.

\section{Conclusion}
We present \OURMODEL, the first video generation model fully trained on NPUs, achieving SOTA performance both on VBench and user studies. By initializing from Stable Diffuison 3.5 Large with minimal architectural changes, implementing a novel phased joint progressive training strategy for simultaneous image-video optimization, and RLHF adaptation for video generation, our framework produces high-quality, diverse videos across varying resolutions and durations while converging in just one month. Remarkably, \OURMODEL~matches the performance of current leading open-source models with only half their parameters, demonstrating unprecedented training efficiency without compromising output fidelity. This work not only advances the frontier of scalable video synthesis but also establishes a new paradigm for large-scale generative model training on alternative hardware architectures.

\section*{Acknowledgements}
We sincerely appreciate our collaborators at ByteDance for their support. They contributed to this work but were not listed as authors. Tao Jiang, Xiong Ke, Xiaotong Shi, Xinjie Huang, Xukai Jiang, Junfu Wang, Zirui Guo, Shuhan Yao, Zhenyu Huang, Hanyu Li, Ziyu Shi, Siqi Wang, Yan Qiu, Yaling Mou, Qingyi Wang, Fengxuan Zhao.

\clearpage
\newpage
\bibliographystyle{plainnat}
\bibliography{neurips_2025}

\end{document}